\documentclass[runningheads]{llncs}

\usepackage{eccv}

\usepackage{eccvabbrv}

\usepackage{multirow}
\usepackage{multicol}
\usepackage{xcolor}
\usepackage{colortbl}

\usepackage{graphicx}
\usepackage{booktabs}

\usepackage[accsupp]{axessibility}  % Improves PDF readability for those with disabilities.

\usepackage[accsupp]{axessibility}  % Improves PDF readability for those with disabilities.

\usepackage{hyperref}

\usepackage{orcidlink}

\begin{document}

\title{
%SAM-COD: A Unified Framework for Weakly Supervised Camouflaged Object Detection\\
%KD-SAM: Weakly Supervised  Camouflaged Object Detection with SAM-based Knowledge Distillation.\\
SAM-COD: SAM-guided Unified Framework for Weakly-Supervised Camouflaged Object Detection}
%%%%%%%%%%%%%%%%%%%%%%%%%%%%%

\titlerunning{SAM-COD}

\author{Huafeng Chen\inst{1,3} \and
Pengxu Wei\inst{2,4}\and
Guangqian Guo\inst{1,3}\and Shan Gao\inst{1,3}\thanks{Corresponding author}}

%Corresponding \dag

\authorrunning{H.~Chen et al.}

\institute{$^{1}\ $Northwestern Polytechnical University, $^{2}\ $Sun Yat-Sen University, $^{3}\ $National Key Laboratory of Unmanned Aerial Vehicle Technology, $^{4}\ $Peng Cheng Laboratory, China 
\\\email{ 
\{\{chf, guogq21\}@mail, gaoshan@\}.nwpu.edu.cn, weipx3@mail.sysu.edu.cn}}

% \\ \url{http://www.springer.com/gp/computer-science/lncs}

\maketitle

\begin{abstract}
%  Existing Camouflaged Object Detection (COD) methods heavily rely on mask labels, but it takes about 60 minutes to annotate an accurate mask, which is time-consuming and labor-intensive. In contrast, weakly supervised labels such as scribble, box, and point require about only 10, 5, and 2 seconds, respectively. 
%  \textcolor{blue}{SAM, as a foundational visual model, exhibits outstanding segmentation capabilities across multiple types of prompts.}
%  Current weakly supervised COD methods have trouble simultaneously supporting all types of labels and exhibit significantly lower performance compared to fully supervised COD methods. 
%  In this paper, we propose a unified framework, SAM-COD, supporting arbitrary weakly supervised labels by integrating Segment Anything Model (SAM). To address the challenges associated with using SAM directly, including 1) Prompt compatibility of scribble, 2) Extreme response, 3) Semantically erroneous response, and 4) Unstable feature representation, we design: a) Prompt adapter, b) Response filter, c) Semantic matcher, and d) Prompt-Adaptive knowledge distillation. On three mainstream COD benchmarks, experimental results show that our model outperforms state-of-the-art fully-supervised methods across various metrics. Our codes will be released soon.

  Most Camouflaged Object Detection (COD) methods heavily rely on mask annotations, which are time-consuming and labor-intensive to acquire. Existing weakly-supervised COD approaches exhibit significantly inferior performance compared to fully-supervised methods and struggle to simultaneously support all the existing types of camouflaged object labels, including scribbles, bounding boxes, and points. Even for Segment Anything Model (SAM), it is still problematic to handle the weakly-supervised COD and it typically encounters challenges of prompt compatibility of the scribble labels, extreme response, semantically erroneous response, and unstable feature representations, producing unsatisfactory results in camouflaged scenes. To mitigate these issues, we propose a unified COD framework in this paper, termed SAM-COD, which is capable of supporting arbitrary weakly-supervised labels. Our SAM-COD employs a prompt adapter to handle scribbles as prompts based on SAM. Meanwhile, we introduce response filter and semantic matcher modules to improve the quality of the masks obtained by SAM under COD prompts. To alleviate the negative impacts of inaccurate mask predictions, a new strategy of prompt-adaptive knowledge distillation is utilized to ensure a reliable feature representation. To validate the effectiveness of our approach, we have conducted extensive empirical experiments on three mainstream COD benchmarks. The results demonstrate the superiority of our method against state-of-the-art weakly-supervised and even fully-supervised methods. 
  % Our source codes and trained models will be publicly released. 

  \keywords{Weakly-Supervised Camouflaged Object Detection \and SAM \and Prompt Adapter \and Prompt-Adaptive Knowledge Distillation}
\end{abstract}

\section{Introduction}
\label{sec:intro}

Camouflaged Object Detection (COD) aims to detect concealed objects from various backgrounds~\cite{fan2020camouflaged,pang2022zoom,zhai2021mutual,zhan2023camouflage,chen2021exploring}, which have imperceptible visual appearances with extremely high similarity to the environment. It holds great promise for practical applications, \emph{e.g.}, species discovery~\cite{perez2012early,fan2020pranet,fan2020inf,ji2021progressively,le2019anabranch,ji2021learning}, medical image segmentation, and animal tracking~\cite{fan2021concealed}. Considering that mask annotations as fully-supervised learning labels~\cite{fan2020camouflaged} are not always available for the time-consuming and laborious cost, \emph{e.g.}, 60 minutes per image~\cite{fan2020camouflaged}, weakly-supervised labels are promising as an attractive alternative, including scribble ($\sim$10 seconds)~\cite{he2023weakly}, bounding box ($\sim$5 seconds), point ($\sim$2 seconds), \emph{etc}. 

\begin{figure}[t]
    \centering
    \includegraphics[width=10cm]{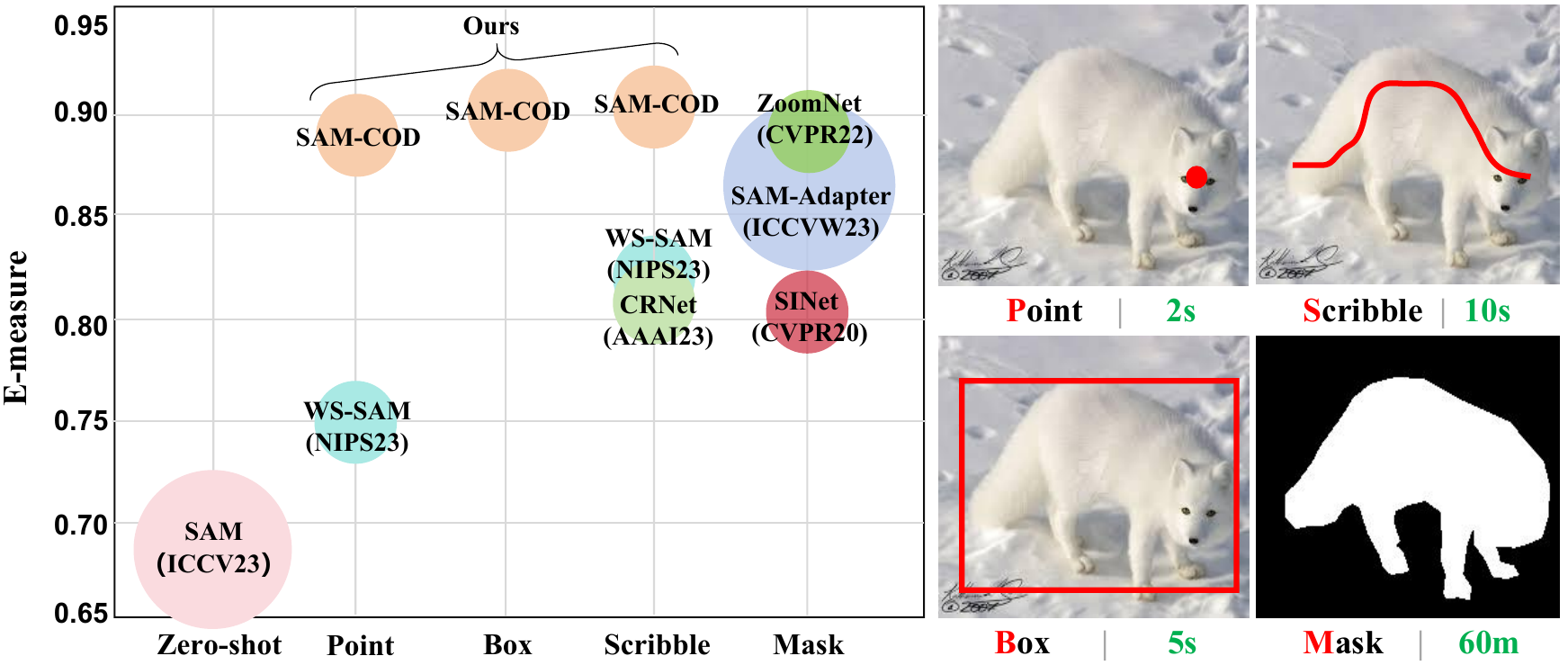}\\
    % \[]{-5pt}
    \caption{Comparison of COD methods for different granularity labels.
    A larger circle denotes a higher-parameter model. SAM-COD is capable of handling three different labels for camouflaged objects. It achieves the highest performance under the weakly-supervised learning setting and even outperforms the fully supervised ZoomNet~\cite{pang2022zoom}. 
    }
    % \[]{-12pt}
    \label{fig:overview}
\end{figure}

However, few works explore how to employ weakly-supervised labels for COD. 
% \textcolor{red}{There is only one work, CRNet~\cite{he2023weakly}, that utilizes scribble annotation to address weakly-supervised COD,} but it has a significant performance gap from fully supervised COD methods. 
There are only two works, CRNet~\cite{he2023weakly} utilizes scribble annotation, and WS-SAM~\cite{he2024weakly} utilizes scribble and point annotation to address weakly-supervised COD. However, they exhibit a significant performance gap compared to fully supervised COD methods.
Thus, in this paper, we make an early attempt to explore a unified resolution of weakly-supervised COD for different weakly-supervised labels, including \emph{point}, \emph{bounding box}, and \emph{scribble}, achieving comparable performance to fully supervised COD methods, shown in Fig.~\ref{fig:overview}.

% For instance, scribble annotation takes just 10 seconds, providing a 360-fold reduction in annotation time compared to mask annotation. The difficulty of weakly-supervised Camouflaged Object Detection (WSCOD) =is substantial, and currently, only CRNet\cite{he2023weakly} has attempted to use scribble annotation to accomplish WSCOD. 
% %However, CRNet has not effectively balanced annotation efficiency and model performance. Its performance only approaches that of early subpar fully supervised models, with a significant gap from SOTA fully supervised COD methods. 
% However, the performance of CRNet has a significant gap from SOTA fully supervised COD methods.
% Therefore, we pondered, “Can WSCOD methods achieve results comparable to fully supervised COD methods?” 

% The emergence of Segment Anything Model (SAM)\cite{kirillov2023segment} provides a potential. Unfortunately, there has been no prior research exploring this avenue. Therefore, we attempt to use SAM to address WSCOD and proposed a unified WSCOD framework, incorporating more time-efficient box and point annotations. To the best of our knowledge, we are the first to utilize box and point labels in the WSCOD task, achieving performance close to the best COD methods, as demonstrated in Fig. 1.
%To the best of our knowledge, we are the first to unify commonly used weak supervision labels, including Point, Box, and Scribble, in the WSCOD task, and have achieved performance surpassing state-of-the-art COD methods, as shown in Figure 1.

\begin{figure}[tb]
\centering
\includegraphics[width=12cm]{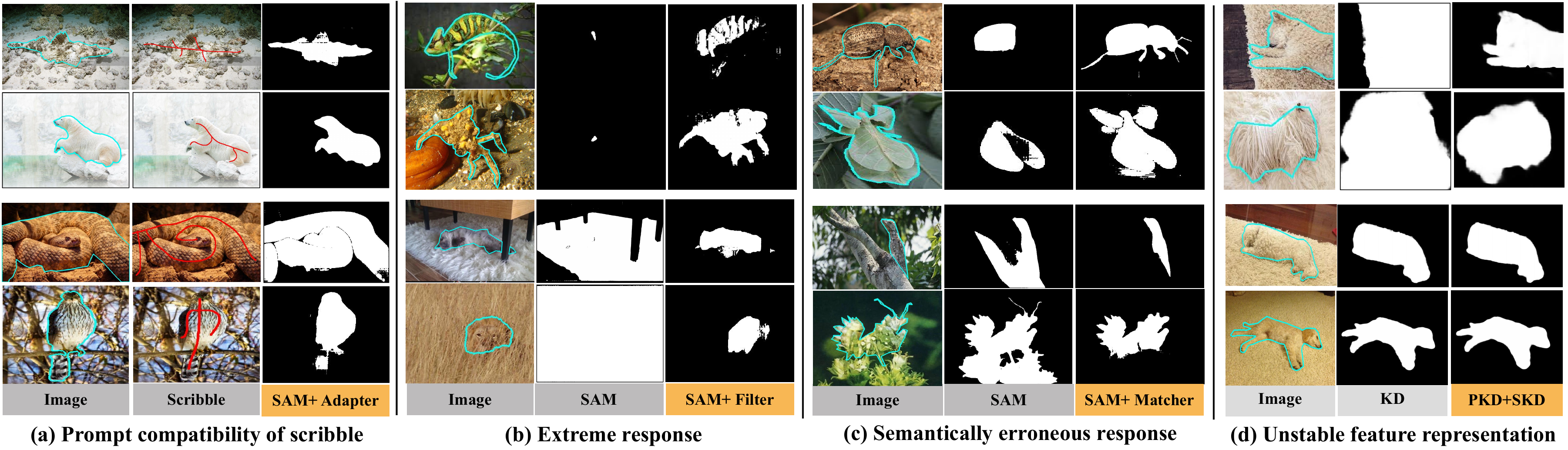}\\
\caption{Issues arising from SAM in COD, \emph{i.e.}, 
a) prompt compatibility of scribble: SAM does not support the scribble input.
%inputting SAM with multiple points, but underperformance (column 2). 
b) extreme response: SAM produces extensive background responses (rows 3, 4) and minimal object responses (rows 1, 2). 
c) semantically erroneous response: SAM produces erroneous responses to non-camouflaged objects (rows 3, 4) and object-biased fine-grained semantic responses (rows 1, 2).  
d) unstable feature representation: SAM produces varied outcomes (1, 2 rows vs. 3, 4 rows) in similar scenarios. 
The contours of camouflaged objects are highlighted in blue.}
\label{fig:SAM-issue}
\end{figure}

Although Segment Anything Model (SAM)~\cite{kirillov2023segment} directly provides candidates for WSCOD, it is not trivial to address WSCOD task with the aid of SAM. It mainly faces four typical challenges,  
\emph{1) Prompt compatibility of scribble:} 
%SAM does not support the crucial scribble-type annotations~\cite{he2023weakly} in WSCOD.
SAM mainly supports box, point, and text-type inputs, but does not support scribble inputs which are applicable for existing WSCOD~\cite{he2023weakly}, as shown in Fig.~\ref{fig:SAM-issue}(a). Then, the direct use of point input does not always yield satisfactory results. It is desirable to explore how to make different types of annotations in WSCOD compatible with SAM.
\emph{2) Extreme response:} 
%\pengxu{Under the prompt,} 
For COD, SAM is prone to producing erroneous responses in extremely small regions or the entire background area, as shown in Fig.~\ref{fig:SAM-issue}(b). This is due to the protective features of camouflaged objects, such as various mimetic patterns, spots, and low-contrast surface textures. 
%usually the camouflaged target has evolved unusual protective features, such as various mimetic patterns, spots, low-contrast boundaries, and body surfaces. Therefore, with limited prompt information, the response of SAM can easily fall into local mimicry (Fig. 2 a) or spread across the entire background area (Fig. 2 b). So we proposed the response Filter to reduce such obvious erroneous responses. 
%
\emph{3) Semantically erroneous response:} SAM is also prone to wrong semantic responses for camouflage objects, including 
%that do not meet the semantic requirements of COD. %(“camouflage”,“whole”). % under the prompt. 
Specifically, a) non-camouflaged object response: SAM lacks training on relevant data, and lacks an understanding of camouflage semantics, b) local response: SAM has a rich segmentation granularity, making it prone to generating local semantic responses, as shown in Fig.~\ref{fig:SAM-issue}(c). 
%\textcolor{red}{
\emph{4) Unstable feature representation:} 
%the Knowledge Distillation (KD) in WSCOD are difficult, 
The images of WSCOD task can exhibit completely different performance in very similar situations, as shown in Fig.~\ref{fig:SAM-issue}(d). 
% 除了WSCOD任务之外,SAM本身也是导致问题之一; a gap between SAM and x;
This is due to COD scenarios being challenging, and there is a significant gap in scale between the foundation model SAM and the student model. Direct distillation with limited supervision results in unstable learned features.

In this work, we propose a unified weakly-supervised COD framework, \textbf{SAM-COD}, supporting the input of arbitrary weakly-supervised label, \emph{i.e.}, point, box or scribble by integrating the large visual model, SAM. We forgo the use of fully supervised labels for fine-tuning SAM and instead explore the use of weakly-supervised labels to prompt SAM. To mitigate the issues aforementioned, we first introduce the Prompt Adapter, which extracts the skeleton of the scribble label and then discretely samples it to points, making it compatible with SAM.
%Hence, we first introduce the Prompt Adapter, which extracts the skeleton of the scribble label and then discretely samples it to points, making it compatible with SAM.
%%%%% BY_CHF:11.18 %%%%
Subsequently, we devise a Response Filter to filter out extreme responses from SAM by computing the ratio of the mask to the image size.
Then, we construct a Semantic Matcher, which measures the semantic score of the mask by semantic entropy, which combines with the segmentation score of SAM to select masks that balance segmentation details and accurate semantics. 
We design a Prompt-Adaptive Knowledge Distillation according to different types of prompts, which enhances knowledge distillation by introducing prompt-guided knowledge for COD tasks, improving the quality of feature representation distilled from SAM.

Overall, our contributions are summarized as follows:
% \[]{-5pt}
\begin{itemize}
% \pengxuc{Contributions are too simple. In my opinion, 1) framework, 2) the design of scribble, regulation, 3) KD, 4) experiments. Please re-organize this part! Important!}
% \item{To the best of our knowledge, this is the first attempt to leverage the vision foundation model to address the COD task in weak annotation.}
% \item{In order to evaluate the model, we relabel the COD dataset, created point label Dataset P-COD and box label Dataset B-COD.}
\item {
We present a novel unified framework inheriting from SAM, integrating three supervision labels, \emph{i.e.}, \emph{scribble}, \emph{bounding box}, and \emph{point}, for weakly-supervised camouflaged object detection. To the best of our knowledge, 
% this is the first WSCOD method to support all the existing weakly-supervised labels.
this is the first WSCOD method to support all current weakly-supervised labels.
}
%It employs a Prompt-Adapter together with scribble skeletonization and discretization, enabling \emph{scribble} to adapt well to SAM-based models.}

\item {We devise Response Filter and Semantic Matcher modules, addressing the issue that SAM is error-prone to producing unreliable extreme responses in COD scenarios, to obtain high-quality object masks.} 
%This is simple, but effective.}

\item{We propose a Prompt-adaptive Knowledge Distillation (PKD) for WSCOD. The distilled knowledge could be adaptively learned according to the three types of input prompts, which promotes knowledge distillation in WSCOD by focusing on distillation in high-value regions within the camouflage scene.}
%\pengxuc{add one explanation for its advantage. Do not claim from prior. I remember you summarized the key idea from the learning of three labels. This would be more insightful.}

% We design a novel knowledge distillation method, Prompt-Adaptive Knowledge Distillation (PKD), specifically for WSCOD. PKD introduces beneficial prior knowledge for COD through prompts, enabling targeted distillation of the model.
\item{We conduct extensive experiments on three widely-used COD datasets, demonstrating that our method achieves state-of-the-art performance. 
To the best of our knowledge, this is the first WSCOD method to outperform the state-of-the-art fully supervised methods under all the weakly-supervised labels.
% and even outperforms the state-of-the-art fully supervised methods only under weakly supervised labels.
Moreover, when migrated to Salient Object Detection (SOD) and Polyp Segmentation tasks, our framework also achieves favorable results.}
%on four widely-used SOD datasets.}
\end{itemize}

% \begin{figure}[t]
% \centering
% \includegraphics[width=12.5cm]{sec/Fig/New.F.333.pdf}\\
% \caption{
% % 应该是main idea ,需要改。
% The architecture of the proposed SAM-COD framework. Prompt Adapter supports scribbles to adapt the input prompt of SAM. Response Filter handles the extreme responses of SAM. Semantic Matcher is utilized to solve SAM's response issues arising from a lack of semantics in COD. Prompt-Adaptive Knowledge Distillation is designed for knowledge distillation in WSCOD.
% }\label{fig:framework}
% \end{figure}

% \[]{-5pt}
\section{Related Work}
% \[]{-5pt}
\textbf{Camouflaged Object Detection.}
COD focuses on detecting camouflaged objects within an image. SINet\ \cite{fan2020camouflaged} proposes a COD dataset with 10K camouflaged images, which takes an average of around 60 minutes to annotate each image.
\cite{mei2021camouflaged, sun2021context} attempt to mine inconspicuous features of camouflage objects from the background through
meticulously designed feature exploration modules.
ZoomNet\ \cite{pang2022zoom} introduces a mixed-scale triplet network to address the challenges posed by COD. The aforementioned COD methods heavily rely on large-scale datasets with pixel-level annotations. However, the unclear boundaries make pixel-wise annotation of camouflaged objects a time-consuming and labor-intensive task. CRNet\ \cite{he2023weakly} was the first to introduce the S-COD dataset, which employs scribble annotations as weak supervision. WS-SAM\ \cite{he2024weakly} employs scribble and point annotations as weak supervision, but there is no dataset constructed with point annotation. Furthermore, box annotation has yet to be explored. So we propose box and point annotations to construct COD datasets.
% , the first box-supervised and point-supervised dataset for COD task to our knowledge. 
Furthermore, we propose the first model that simultaneously supports various weak supervision labels and outperforms fully supervised methods. 

%Instead, points are randomly sampled from training masks for point annotation. 
%But scribble annotations are not the most time-efficient form of weak supervision, and more time-saving methods such as box 

%Our model firstly support all type of weak annotations.
%We investigate the application of scribble, box, and point annotations in WSCOD and unify them with the help of SAM.

\noindent\textbf{SAM in COD.} SAM \cite{kirillov2023segment} excels in traditional segmentation tasks, achieving remarkable results, sometimes matching the performance of fully supervised methods, even in a zero-shot setting. \cite{chen2023sam, tang2023can} indicate that while SAM shows promise in generic object segmentation, its performance on the COD task is constrained. SAM-Adapter\ \cite{chen2023sam} employs an adapter for efficient tuning instead of relying on traditional fine-tuning methods. This adaptation enables SAM to align with the data distribution in COD, reducing the cost of fine-tuning while simultaneously enhancing the performance of SAM in COD. WS-SAM~\cite{he2024weakly} processes three augmented images through SAM and fuses the obtained masks to obtain the final pseudo-label. But the drawbacks of it are also obvious: 1) tripled SAM inference time 2) the full potential of SAM was not utilized, and only the highest scoring mask was used instead of the top-3 masks. We apply SAM to design a unified framework for point, box, and scribble annotations.
%In contrast to prior work, we apply SAM to WSCOD task and design a unified framework for point, box and scribble annotations.

\noindent\textbf{Knowledge Distillation.}
Knowledge distillation (KD)\ \cite{buciluǎ2006model,hinton2015distilling} has been primarily designed to train a small network to mimic the
output of a larger network to compress models. DINO\ \cite{caron2021emerging} has introduced a straightforward self-supervised method, which can be described as a label-less self-distillation model to optimize the representation learning. Distillation under WSCOD differs from traditional distillation, as 1) the COD scenario is challenging, and 2) there is little supervision. This makes traditional distillation methods unsuitable, and currently, there is a lack of exploration into distillation under the WSCOD task. So we design the prompt-adaptive knowledge distillation for the WSCOD task.
%which is compatible with different kinds of input prompts.}
%Unlike traditional distillation, we design the prompt-adaptive knowledge distillation for the WSCOD task, which is compatible with different kinds of input prompts.

%WSCOD+SAM本身在蒸馏就不常见
%目前缺少特殊场景下->SKD;
%弱监督条件下：->PKD

% \section{Preliminaries}
% %主要是三个Score的
% The SAM model provides an interface for Prompts, supporting different types of Prompts such as Point, Box, and text. Especially for limited cues related to multiple objects, SAM can output multiple valid object masks along with associated confidence scores.
% \begin{equation}
%     (S_{1},V_{1};S_{2},V_{2};S_{3},V_{3})=\text{SAM}(I,P),
% \end{equation}
% where $V_{i}$ donate the i-$th$ objects masks and $S_{i}$ represents the corresponding confidence score, $P$ donate prompt.

\begin{figure}[t]
\centering
\includegraphics[width=12.5cm]{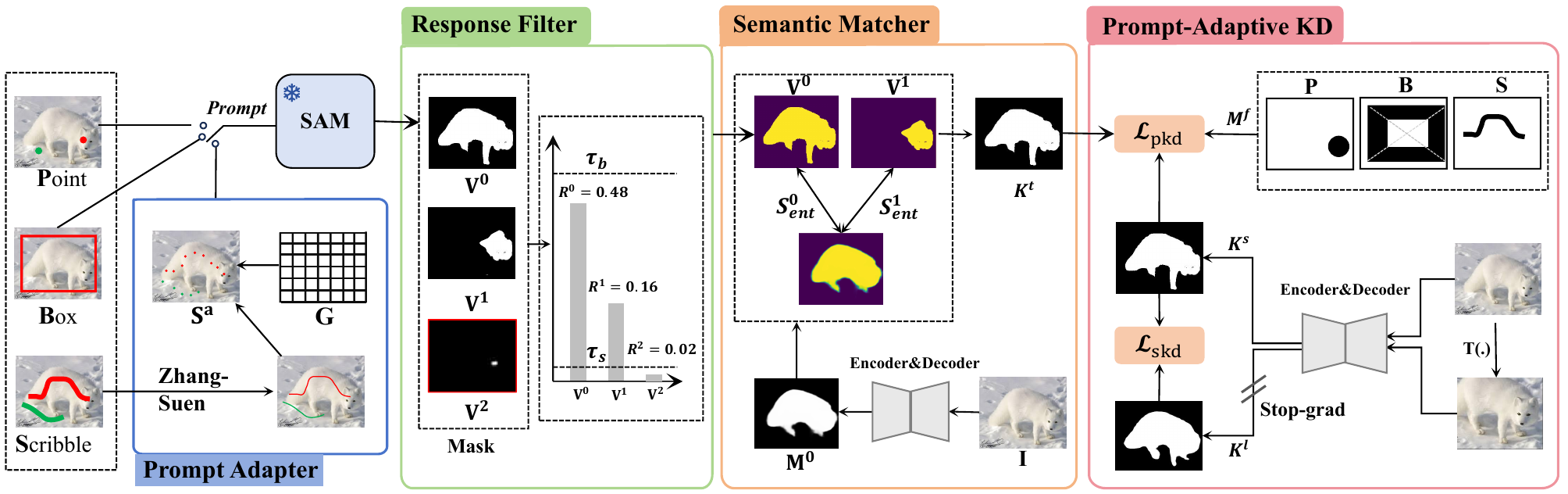}\\
\caption{
% 应该是main idea ,需要改。
The architecture of the proposed SAM-COD framework. Prompt Adapter supports scribbles to adapt the input prompt of SAM. Response Filter handles the extreme responses of SAM. Semantic Matcher is utilized to solve SAM's response issues arising from a lack of semantics in COD. Prompt-Adaptive Knowledge Distillation is designed for knowledge distillation in WSCOD.
}\label{fig:framework}
\end{figure}

\section{Approach}

%\textbf{Framework.}
The overall architecture of the proposed framework is shown in Fig.~\ref{fig:framework}. 
%a SAM with frozen parameters, used to obtain masks, a segmentation model composed of an encoder and a decoder that requires training for parameter updates. 
Prompt adapter is used to process scribbles to adapt SAM prompt input. Response filter is employed to handle extreme response situations of SAM under the prompt. Semantic matcher is utilized to improve SAM's response issues arising from a lack of COD-related semantics. Prompt-adaptive knowledge distillation is employed for the knowledge distillation in WSCOD. %Details can be found in the following parts.

%缺乏每个组件一句话的描述：
%The overall architecture of the proposal framework is shown in Fig. 3, which consists of the teacher model for knowledge transfer and the student model for knowledge distillation. The teacher model primarily includes SAM, a) Prompt Adapter, b) Attention Regulator, and c) Pattern Matcher. The student model mainly encompasses distillation and self-distillation.

%Unified labels as prompt

\subsection{Prompt Adapter}
%我们整合Weakly Supervised Label作为prompt，以利用SAM开放的口，具体如下所示：
We use three kinds of weakly-supervised labels as prompts: point, box, and scribble. SAM directly supports types of point and box as input prompts. Unfortunately, SAM does not support scribble-type prompt. Therefore, we design a prompt adapter to convert scribbles into discrete points, making it compatible with SAM, as shown in Fig. \ref{fig:framework}.

Specifically, 
%as shown in Fig. 3(a), 
we first use the Zhang-Suen algorithm\ \cite{zhang1984fast} to extract the skeleton of the scribble. Then, we perform discrete sampling on it. Specifically, we first create a grid G, where the grid points are uniformly distributed and the distance is $min(\alpha W, \alpha H)$, where $H$ and $W$ represent the length and width of the input image, respectively. $\alpha$ is the hyperparameter. Afterwards, we form a discrete point set $S^a$ by sampling points that coincide with both the scribble skeleton and grid lines. By now, we obtain the SAM prompt: $prt=\{P,B,S^a \}$, where $P$ and $B$ indicate Point and Box labels, respectively. 
\subsection{Response Filter} 
In COD, the camouflage objects usually exhibit excellent mimicry. So, SAM is prone to locate the extreme response under limited prompts, as shown in Fig.~\ref{fig:SAM-issue}(b). To solve this, we design a response filter to prevent taking advantage of these evidently abnormal responses, as shown in Fig. \ref{fig:framework}.

Specifically, SAM outputs three valid masks and corresponding confidence scores given the prompt input:
\begin{equation}
\{V^{i},S_{con}^{i}|i=1,2,3\}={SAM}(I,prt),
\end{equation}
where $V^{i}$ donate the $i$-$th$ objects masks and $S_{con}^{i}$ represents the corresponding segmentation confidence score. SAM defaults to using the mask with the maximum confidence score. 
Subsequently, we design a response filter to determine whether the mask exhibits extreme response by calculating the ratio of the mask size to the image size:
\begin{equation}
    R^{i} =\mathbb{I}(\tau_{s}<\frac{A^{i}}{HW}<\tau_{b}),
\end{equation}
where $\mathbb{I}(\cdot)$ is an indicator function. $A^{i}$ is the area of the $i$-th mask $V^i$. $\tau_{s}$ and $\tau_{b}$ represent the maximum and minimum thresholds, respectively. 

\subsection{Semantic Matcher} 
SAM lacks the semantic knowledge,
%required for COD tasks, 
specifically the semantic understanding of camouflaged and overall granularity,
%specifically the semantic understanding of camouflaged objects and the overall semantic segmentation of the target (semantic granularity), 
leading to responses that do not align with the objects, as shown in Fig.~\ref{fig:SAM-issue}(c). To solve it, we design a semantic matcher to measure the semantic score by the semantic entropy. It then 
selects masks with accurate semantics, as shown in Fig. \ref{fig:framework}.
%combines with SAM's segmentation confidence score, selecting masks that balance both segmentation quality and accurate semantics.

Specifically, we first train the model on COD data to obtain the mask $M^o$: 
\begin{equation}
    M^{o} =D(E(I)),
\end{equation}
where $I$ donates input image, $E$ and $D$ are the encoder and decoder of the model, respectively. Although $M^o$ may not rival the masks of SAM in segmentation details, training on COD data %makes the model include semantic of overall and camouflage.
provides the model with a preliminary understanding of camouflage semantics.
\iffalse
Then, we design semantic entropy $SE^i$ using $M^o$ to measure the semantic score of the mask $V^i$:
\begin{equation}
    SE^i =-\sum_{j}{M^o_jlog(V^i_j)+(1-M^o_j)log(1-V^i_j)},
\end{equation}
where $j$ is the pixel index. Smaller values of $SE^i$ indicate higher semantic score for $V^i$. So we design semantic score $S_{ent}^i$ as:
\begin{equation}
    S_{ent}^i=\frac{1}{SE^i}
\end{equation}
%\textcolor{red}{Ultimately, we select the mask $V^i$ with the the maximum ratio of $SC^i$ to $SE^i$, which balancing segmentation details and accurate semantics, as the final knowledge distillation source $K^{t}$.}
Ultimately, we select the mask with the highest product of $S_{ent}^i$ and $S_{con}^i$ scores, which balance segmentation details and accurate semantics, forming the optimal $V^i$ as:
 %the final distillation source $K^{t}$:
 \begin{equation}
    V^{opt}=V^i,\quad \text{where} \quad i=index(max(S_{ent}^iS_{con}^i)),
 \end{equation}
 \fi

%%%%%%%%%%%%%%%%%%%%%% GS reformulate %%%%%%%%%%%%%%%%%%%%%%%%%%%%

Then, we design semantic entropy $S_{ent}^i$ using $M^o$ to measure the semantic score of the mask $V^i$:
\begin{equation}
    S_{ent}^i =-\sum_{j}{M^o_jlog(V^i_j)+(1-M^o_j)log(1-V^i_j)},
\end{equation}
where $j$ is the pixel index. Smaller values of $S_{ent}^i$ indicate higher semantic score for $V^i$. 

We select the mask with the highest product of $\frac{1}{S_{ent}^i}$ and $S_{con}^i$ scores, which balance segmentation details and accurate semantics, forming the optimal mask $V^{opt}$ in $V^i$ as:
 %the final distillation source $K^{t}$:
 \begin{equation}
    opt=argmax(\frac{S_{con}^i}{S_{ent}^i}).
 \end{equation}
%%%%%%%%%%%%%%%%%%%%%%%%%%%%%%%%%%%%%%%%%%%%%%%%%%%%%%

\subsection{Prompt-Adaptive Knowledge Distillation}
%\textbf{Prompt Knowledge-distillation.} 
%The knowledge distillation in WSCOD differs from traditional tasks. 
%We forgo the use of fully supervised labels for fine-tuning SAM and instead explore the use of weaker labels to prompt SAM, reducing data costs. 
We employ knowledge distillation method to transfer the knowledge from large visual model SAM to a smaller model, thereby reducing the data cost and model size. 
%3) During the distillation phase, we decouple SAM from the model, freezing SAM while employing offline forward propagation for SAM, i.e., pre-extracting knowledge information from SAM's output to reduce computational demands.
However, COD task is challenging and the weak supervision makes knowledge distillation more difficult.
%if we want distillate knowledge from a large model to a small one.
Specifically, the proposed framework distillates the optimal mask $V^{opt}$ from SAM as \textbf{t}eacher knowledge $K^t$ to the \textbf{s}tudent knowledge $K^s$ in our model. Moreover, we leverage the prior knowledge of different prompts to enhance the distillation quality, 
%namely Prompt-Adapter Knowledge Distillation (PKD).

\noindent{\textbf{Prompt-adaptive Mask Generation}}. The input prompts (scribble, box, and point) contain the structure, boundary, and discriminative region of camouflaged objects, respectively. 
%我们设计了Prompt-Adaptive-to-attention mask, 使得蒸馏过程更关注那些更重要的区域，生成attention weight mask M^f的过程如下，二值分割Box和Point,边缘信息是二分利用从Box到中心，分为边缘区和内部区。同理Point也是以P-Mask sum完像素数目，开根号代表大小，把大小分成两半，远端是不明显区域，近端是明显区域。
%(“points label” in low camouflage regions)
%%%% Adding in 11.16 %%%%
These have been confirmed to be crucial for COD tasks~\cite{he2023weakly,zhan2023camouflage}. Therefore, we construct a prompt-adaptive mask $M^f$ for the knowledge distillation according to the input prompt. The key distillation regions in $M^f$ are marked as 0 (black areas).
Specifically, 1) Scribble label, retaining the labeled foreground while discarding the background yields the corresponding $M^f$. 
% 2) Point label, a circle area with the radius of $R$ in Eq.\ 2. 
2) Point label, an inscribed circle of $K^t$ with point label as the center.
%appropriate outward expansion serves as the discrimination region. Specifically, we adjust  with the Point as the center to obtain the inscribed circle of
3) Box labels, represented by \textit{bold boxes}, with edge width and height being one-fourth of the length and width of box label, respectively.
%3) Box label, representing as a \textit{bold box}
%$(x1, y1, x2, y2)$, are divided into two parts. The rectangular region near the center $((3x1+x2)/4,(3y1+y2)/4,(3x2+x1)/4,(3y2+y1)/4)$ is discarded as the non-edge region, and the remaining annular region is used as the corresponding $M^f$. 
%$K^t$, which becomes the corresponding $M^f$.

% 

Then, the prompt-adaptive knowledge distillation loss is defined as:
\begin{equation}
 \mathcal{L}_{pkd}=-\sum_{j} M^F ( K^t_jlog(K^s_j)+(1-K^t_j)log(1-K^s_j)),
\end{equation}
%-\sum_{i\in M^f}K^t_ilog(K^s_i)+(1-K^t_i)log(1-K^s_i)
where $K^s$ is the prediction mask and $j$ is the pixel index. $M^F$$=$$1$$+$$\mathbb{I}$($M^f$$=$0) and $\mathbb{I}(\cdot)$ is an indicator function. $M^F$ as a coefficient in the distillation loss to allocate weight to prompt-guided regions, guides the distillation process to focus on learning key distillation regions.

\noindent\textbf{Self-Knowledge Distillation.}  
The learned feature representation of the model may not be robust enough, as shown in Fig.~\ref{fig:SAM-issue}(d). Inspired by Self-Knowledge Distillation (SKD), we design a student model to enhance the representation learning.
Specifically, for image $I$, we adopt visual transformations $T$, selecting from scale, colorjitter, etc. These visual transformations are able to change the appearance of images, as
%\begin{equation} 
   $ I^{t}=T(I)$.
%\end{equation} 

Then we encode and decode the augment images $I^{t}$, and transform them into two prediction maps $K^s$ and $K^{l}$,  
denoting as:
\begin{equation}
K^s=D(E(I)),K^{l}=D(E(I^{t})),     
\end{equation}
%\textcolor{red}{meaning of $K^{SS}$}
%where the prediction map $K^s$, the local prediction map $K^{l}$, the 
Our objective is to minimize the distance between two prediction maps:
\begin{equation} 
   \min \mathcal{D}(K^s,K^{l})=\sum_{i}|K^s_i - K^{l}_i|,
\end{equation} 
where $i$ is the pixel index, when a transformation $T$ (e.g., scale, crop) is applied to the image $I$, this transformation $T$ should be applied to $K^s$ to be aligned with $K^l$. Here we follow the design of SKD, \emph{i.e.}, stopping the gradient ($stopgrad$) update at one end, 
so the SKD loss function is defined as
\begin{equation} 
    \mathcal{L}_{skd}=\mathcal{D}(K^s,stopgrad(K^{l})).\label{eq:8}
\end{equation} 

\iffalse
\begin{algorithm}[t]
    \caption{SKD Pseudocode, PyTorch-like}
    \label{alg:algorithm}
    \begin{algorithmic}
        % \STATE \textcolor{teal}{\# $E$:  backbone} 
        % \STATE \textcolor{teal}{\# $P$:  prediction}
        \STATE \textcolor{magenta}{for}  $I$  \textcolor{magenta}{in} loader: \textcolor{teal}{\# load a minibatch $I$ with n samples}    
        \STATE\mbox\quad $I^1$, $I^2$ = $T_{1}(I)$, $T_{2}(I)$ \textcolor{teal}{\# Different augmentations}
        %\STATE\mbox\quad
        \STATE\mbox\quad $P^1$ = $g(f(I^1))$ \textcolor{teal}{\# Encoder\& Prediction}
        \STATE\mbox\quad $P^2$ = $f(I^2)$ \textcolor{teal}{\# Encoder}
        \STATE\mbox\quad $l$ = $\mathcal{D}(P^1,P^2)$ \textcolor{teal}{\# Loss}
        \STATE\mbox\quad $l.backward()$ \textcolor{teal}{\# back-propagate}        
        \STATE\mbox\quad $update(f, g)$\textcolor{teal}{\# SGD update}
         \STATE
        \STATE \textcolor{magenta}{def} $\mathcal{D}$($P^1$, $P^2$): \textcolor{teal}{\# negative similarity}
        \STATE\mbox\quad $P^2$= $P^2.detach()$\textcolor{teal}{\#  stop gradient}
        \STATE\mbox\quad \textcolor{magenta}{return} $L1(P^1, P^2)$
    \end{algorithmic}\label{A:1}
\end{algorithm}
\fi

A robust feature representation could be learned from the teacher model to the student one by minimizing the above loss.
%, prediction gap using different transformations will be narrowed.
%In other words, the ineffective representations conditioned on a specific transformation will be guided by better representations learned based on other transformations,
%making the learned feature more robust and invariant, being able to better highlight the foreground objects. 

\subsection{Network}
\noindent\textbf{Encoder\&Decoder.} Encoder and decoder designs can be flexibly replaced with existing models.
In this work, we employ PVT~\cite{wang2021pyramid} as the encoder, which obtains multi-scale features ($F_{eat_1}$, $F_{eat_2}$, $F_{eat_3}$, $F_{eat_4}$). The decoder consists of four 3x3 convolutional layers to reduce the channel dimension of $F_{eat_i}$ to $64$, followed by upsampling these $F_{eat_i}$ to the same size. Subsequently, they are combined through concatenation, and finally, a $3\times3$ convolutional layer is used to obtain the final mask. In our method, all encoders and decoders refer to the same model.

\noindent\textbf{Training Details.} Our training process consists of two main steps. In Training Step 1, we train the encoder and decoder in the semantic matcher to obtain the distillation source $K^t$ at the end.
% details are in the supplementary materials. 
In Training Step 2, we use $K^t$ for knowledge distillation to retrain the encoder and decoder. Further details are in the S.M.

\noindent\textbf{Loss.} Compared to other weakly-supervised methods \cite{he2023weakly,yu2021structure,zhang2020weakly}, we have only two losses. The final loss $\mathcal{L}$ includes $\mathcal{L}_{pkd}$ and $\mathcal{L}_{skd}$ defined as:
\begin{equation}
   \mathcal{L}=\mathcal{L}_{pkd}+\mathcal{L}_{skd}.
\end{equation}

\section{Experiments}

\begin{table}[tb]
  \begin{center}
   \caption{Quantitative comparison with state-of-the-arts on three benchmarks. “F”, “U”, “S”, “P”, “Mix” denote fully-supervised label, unsupervised, scribble, point, and mixed random selection of one of three weakly-supervised labels, respectively. “–” is not available. \textcolor{red}{Red} and  \textcolor{blue}{blue} represent the first and second best performance, respectively. }\label{tab:1}
   % \[]{-10pt}
   \renewcommand\arraystretch{1.3}
   \tabcolsep=0.025cm
    {\scriptsize
\begin{tabular}{l|c|cccc|cccc|cccc}
%\toprule
\toprule[0.8pt]
\multirow{2}{*}{Methods}&\multirow{2}{*}{Label}&\multicolumn{4}{c|}{CAMO}&\multicolumn{4}{c|}{COD10K} &\multicolumn{4}{c}{NC4K }\\%&\multicolumn{2}{|c}{ILSVRC2012}\\

&&MAE $\downarrow$&S$_{m} \uparrow$&E$_{m} \uparrow$&F$^{w}_{\beta} \uparrow$&MAE $\downarrow$&S$_{m} \uparrow$&E$_{m} \uparrow$&F$^{w}_{\beta} \uparrow$&MAE $\downarrow$&S$_{m} \uparrow$&E$_{m} \uparrow$&F$^{w}_{\beta} \uparrow$\\

\toprule[0.8pt]
%\emph{Fully-supervised methods}\\
% EGNet $_{\text{\textcolor{gray}{[ICCV19]}}}$ \cite{zhao2019egnet}&F&0.109&0.732&0.800&0.604&0.061&0.736&0.810&0.517&0.077&0.767&0.793&0.626\\
%F3Net $_{\text{\textcolor{gray}{[AAAI20]}}}$\cite{wei2020f3net}&F&$352^2$&-&-&0.109&0.711&0.741&0.564&0.047&0.848&0.894&0.744&0.051&0.739&0.795&0.544&0.069&0.782&0.825&0.706\\
% CSNet $_{\text{\textcolor{gray}{[ECCV20]}}}$ \cite{gao2020highly}&F&0.092&0.771&0.795&0.641&0.047&0.778&0.809&0.569&0.061&0.819&0.845&0.748\\
% ITSD $_{\text{\textcolor{gray}{[CVPR20]}}}$ \cite{zhou2020interactive}&F&0.102&0.750&0.779&0.610&0.051&0.767&0.808&0.557&0.064&0.811&0.845&0.729\\
% MINet $_{\text{\textcolor{gray}{[CVPR20]}}}$ \cite{pang2020multi}&F&0.090&0.748&0.791&0.637&0.042&0.77&0.832&0.608&-&-&-&-\\
% PraNet $_{\text{\textcolor{gray}{[MICCAI20]}}}$ \cite{fan2020pranet}&F&0.094&0.769&0.825&0.663&0.045&0.789&0.861&0.629&-&-&-&-\\
% UCNet $_{\text{\textcolor{gray}{[CVPR20]}}}$ \cite{zhang2020uc}&F&0.094&0.739&0.787&0.640&0.042&0.776&0.857&0.633&0.055&0.813&0.872&0.777\\
SINet~\cite{fan2020camouflaged}&F&0.092&0.745&0.804&0.644&0.043&0.776&0.864&0.631&0.058&0.808&0.871&0.723\\
MGL-R~\cite{zhai2021mutual}&F&0.088&0.775&0.812&0.673&0.035&0.814&0.851&0.666&0.052&0.833&0.867&0.740\\
PFNet~\cite{mei2021camouflaged}&F&0.085&0.782&0.841&0.695&0.040&0.800&0.877&0.660&0.053&0.829&0.887&0.745\\
UGTR~\cite{yang2021uncertainty}&F&0.086&0.784&0.822&0.684&0.036&0.817&0.852&0.666&0.052&0.839&0.874&0.747\\
UJSC~ \cite{li2021uncertainty}&F&0.073&0.800&0.859&0.728&0.035&0.809&0.884&0.684&\textbf{\color{blue}0.047}&\textbf{\color{blue}0.842}&\textbf{\color{red}0.898}&\textbf{\color{blue}0.771}\\

ZoomNet~\cite{pang2022zoom}&F&\textbf{\color{red}0.066}&\textbf{\color{blue}0.820}&\textbf{\color{red}0.892}&\textbf{\color{blue}0.752}&\textbf{\color{blue}0.029}&\textbf{\color{blue}0.838}&\textbf{\color{blue}0.911}&\textbf{\color{blue}0.729}&\textbf{\color{red}0.043}&\textbf{\color{red}0.853}&\textbf{\color{blue}0.896}&\textbf{\color{red}0.784}\\
SAM-Ada.~\cite{chen2023sam}&F&\textbf{\color{blue}0.070}&\textbf{\color{red}0.847}&\textbf{\color{blue}0.873}&\textbf{\color{red}0.765}&\textbf{\color{red}0.025}&\textbf{\color{red}0.883}&\textbf{\color{red}0.918}&\textbf{\color{red}0.801}&-&-&-&-\\
\hline
%\emph{Self\&Weakly-supervised methods}\\
SAM~\cite{kirillov2023segment}&-&0.132&0.684&0.687&0.606&0.050&0.783&0.798&0.701&0.078&0.767&0.776&0.696\\
% DUSD~\cite{zhang2018deep}&U&0.166&0.551&0.594&0.308&0.107&0.580&0.646&0.276&-&-&-&-\\
% USPS~\cite{nguyen2019deepusps}&U&0.207&0.568&0.641&0.399&0.196&0.519&0.536&0.265&-&-&-&-\\
% SS~\cite{zhang2020weakly}&S&0.118&0.696&0.786&0.562&0.071&0.684&0.770&0.461&-&-&-&-\\
SCSOD~\cite{yu2021structure}&S&0.102&0.713&0.795&0.618&0.055&0.710&0.805&0.546&-&-&-&-\\
CRNet~\cite{he2023weakly}&S&0.092&0.735&0.815&0.641&0.049&0.733&0.832&0.576&0.063&0.775&0.855&0.688\\
SAM-S~\cite{kirillov2023segment}&S&0.105&0.731&0.774&-&0.046&0.772&0.828&-&0.071&0.763&0.832&-\\
WS-SAM~\cite{he2024weakly}&S&0.092&0.759&0.818&-&0.038&0.803&0.878&-&0.052&0.829&0.886&-\\
SAM-P~\cite{kirillov2023segment}&P&0.123&0.677&0.693&-&0.069&0.765&0.796&-&0.082&0.776&0.786&-\\
WS-SAM~\cite{he2024weakly}&P&0.102&0.718&0.757&-&0.039&0.790&0.856&-&0.057&0.813&0.859&-\\
\hline
\rowcolor{lightgray!40}SAM-COD&S&\textbf{\color{blue}0.060}&0.836&\textbf{\color{blue}0.903}&0.779&\textbf{\color{blue}0.029}&0.833&\textbf{\color{blue}0.904}&\textbf{\color{blue}0.728}&0.039&0.859&0.912&0.795\\
\rowcolor{lightgray!40}SAM-COD&B&0.062&\textbf{\color{blue}0.837}&0.901&\textbf{\color{red}0.786}&\textbf{\color{red}0.028}&\textbf{\color{red}0.842}&\textbf{\color{red}0.914}&\textbf{\color{red}0.745}&\textbf{\color{red}0.037}&\textbf{\color{red}0.867}&\textbf{\color{red}0.923}&\textbf{\color{red}0.813}\\
\rowcolor{lightgray!40}SAM-COD&P&0.066&0.820&0.885&0.760&0.031&0.831&0.901&0.725&0.041&0.858&\textbf{\color{blue}0.918}&\textbf{\color{blue}0.802}\\

\rowcolor{lightgray!40}SAM-COD &Mix&\textbf{\color{red}0.058}&\textbf{\color{red}0.839}&\textbf{\color{red}0.907}&\textbf{\color{blue}0.784}&0.031&\textbf{\color{blue}0.833}&0.903&0.725&\textbf{\color{blue}0.039}&\textbf{\color{blue}0.862}&0.912&0.798\\
\toprule[0.8pt]
\end{tabular}
}
\end{center}
% \[]{-10pt}
\end{table}

\subsection{Experimental Setup}
\textbf{Datasets.} Our experiments are conducted on three COD benchmarks, CAMO\ \cite{le2019anabranch}, %CHAMELEON\ \cite{skurowski2018animal}, 
COD10K\ \cite{fan2020camouflaged}, and NC4K\ \cite{lv2021simultaneously}. In order to evaluate our method, we first train our network on scribble annotated dataset S-COD\ \cite{he2023weakly}. Subsequently, we re-annotated 4040 images (3040 from COD10K, 1000 from CAMO) to create point-supervised dataset (P-COD) and bounding box-supervised dataset (B-COD) for training, while the remaining images are used for testing. 

%To minimize the labeling time consumption while providing location information of Camouflaged objects. Following previous studies\cite{he2023weakly,fan2020camouflaged}, we relabel 4040 images(3040 from COD10K,1000 from CAMO) and propose the Point-supervised Dataset(P-COD) for training. The remaining is for testing. In Fig.1, we show an example of point annotation and compare it with other stronger annotation methods. For each Camoufaged objects we simulation of the hunting process to choose the discrimative part for labeling, not need to care about the ambiguous boundaries, which has two benefits way, First, easy and natural labeling way, Secondly, partial semantic information is implicitly conveyed to the model through human-simulated prey labeling.
\noindent\textbf{Evaluation Metrics.} We adopt four evaluation metrics: Mean Absolute Error (MAE), S-measure (S$_{m}$) \cite{fan2017structure}, E-measure (E$_{m}$) \cite{ijcai2018p97}, weighted F-measure (F$^{w}_{\beta}$) \cite{margolin2014evaluate}.

\noindent\textbf{Implementation Details.} We implement our method with PyTorch and conduct experiments on one GeForce RTX4090 GPU and use ViT-H version of SAM. We chose PVT-B4~\cite{wang2021pyramid} as the encoder.
%In view of the previous work\cite{kajiura2021improving,he2023weakly}, most set a different resize the image during training and inference,but they are between in 320 and 512. So we test our work in 320 and 512. The batch size is 16 and 8 when resize is 320 and 512. And training epoch is 60. It takes around 4 hours to train in 320 and 7 hours in 512. 
We use the stochastic gradient descent optimizer with a momentum of $0.9$, a weight decay of $5e$-$4$, and triangle learning rate schedule with maximum learning rate $1e$-$3$. The batch size is $8$, and the training epoch is $60$. Input images are resized to $512\times512$. We adopt the offline distillation, 
%where SAM is pre-computed, 
and the forward computation is performed only once, taking only $7$ hours in training. 
%By default, we set $\alpha$=$4$, $\tau$=$150$, and $w$=$15$.

%. In the training phase, input images are resized to $512\times512$. 
%As for the inference process, input images are only resized to $512 \times 512$ and then feed it to our network to predict final maps. 

\begin{figure}[t]
\centering
\includegraphics[width=10cm]{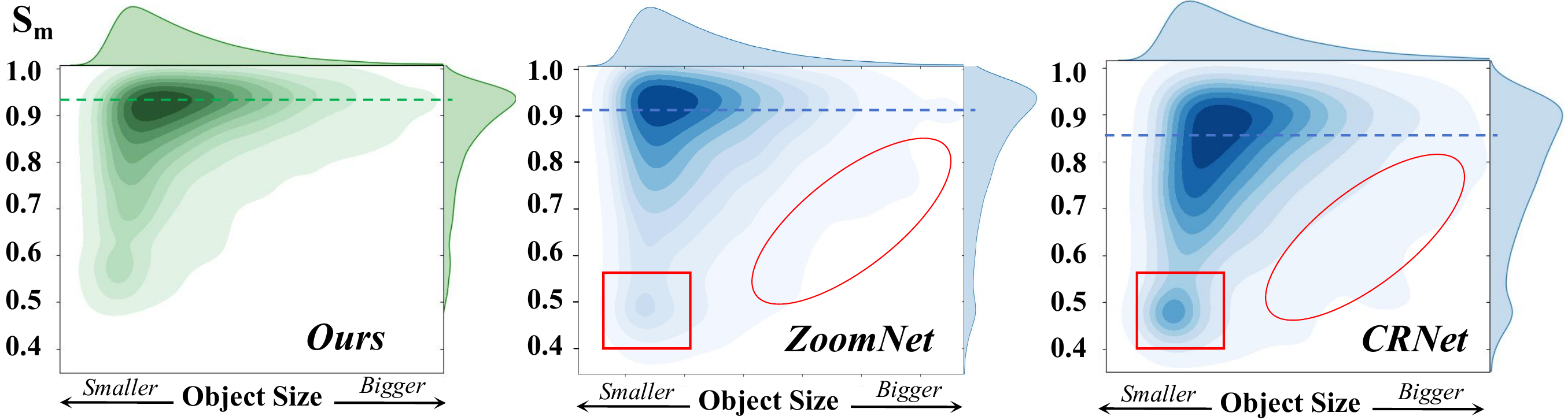}\\
\caption{Density distribution map about $S_m$ and object size. Box and ellipse respectively represent challenging small and big objects, which have poor performance.}
\label{fig:distru}
% \[]{3mm}
\end{figure}

\begin{figure}[th]
\centering
\includegraphics[width=10cm]{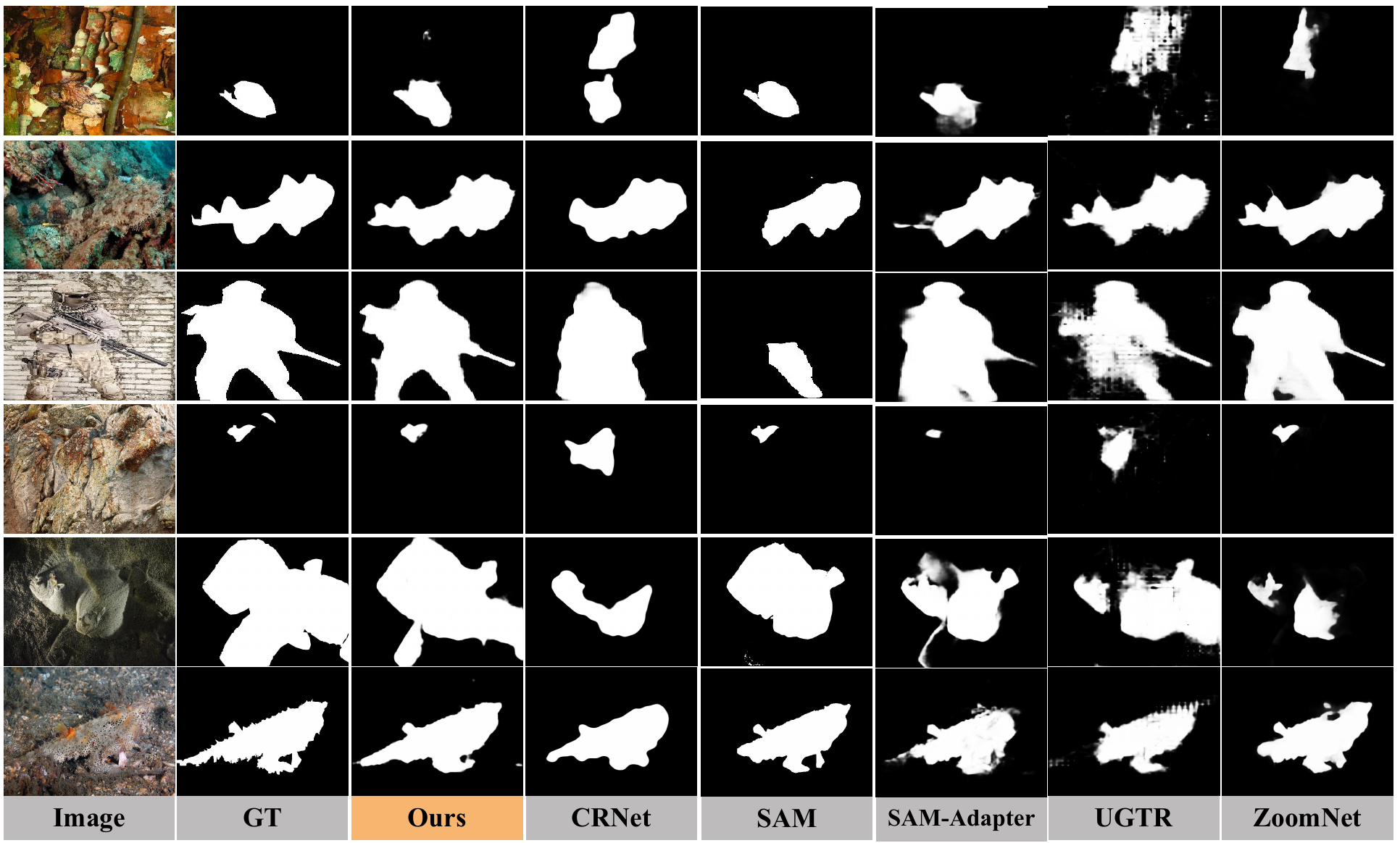}\\%Fig4.pdf;New.F.5.pdf
\caption{Visual comparison with some representative state-of-the-art fully-supervised and scribble-supervised models. }
\label{fig:5}
% \[]{3mm}
% \[]{-10pt}
\end{figure}

\begin{table}[t]
  \begin{center}
  \caption{Comparison of parameters and MACs. “W” denotes the average of three weakly-supervised labels. All metrics are averages of the three datasets.}\label{tab:2}
  % \[]{-10pt}
   \tabcolsep=0.30cm
    {\scriptsize
\begin{tabular}{c|c|cc|cccc}
\toprule[0.8pt]
{Methods}&Label&Para.&MACs&MAE$\downarrow$&S$_{m}\uparrow$&E$_{m}\uparrow$&F$^{w}_{\beta}\uparrow$\\
\toprule[0.8pt]
ZoomNet&F&\textbf{32.38}&95.50&0.046&0.837&0.899&0.755
\\
\rowcolor{lightgray!40}Ours&W&62.64&\textbf{52.63}&\textbf{0.044}&\textbf{0.843}&\textbf{0.907}&\textbf{0.770}
\\
\toprule[0.8pt]
\end{tabular}
}
\end{center}
% \[]{3mm}
% \[]{-10pt}
\end{table}

\begin{figure}[th]
\centering
\includegraphics[width=10cm]{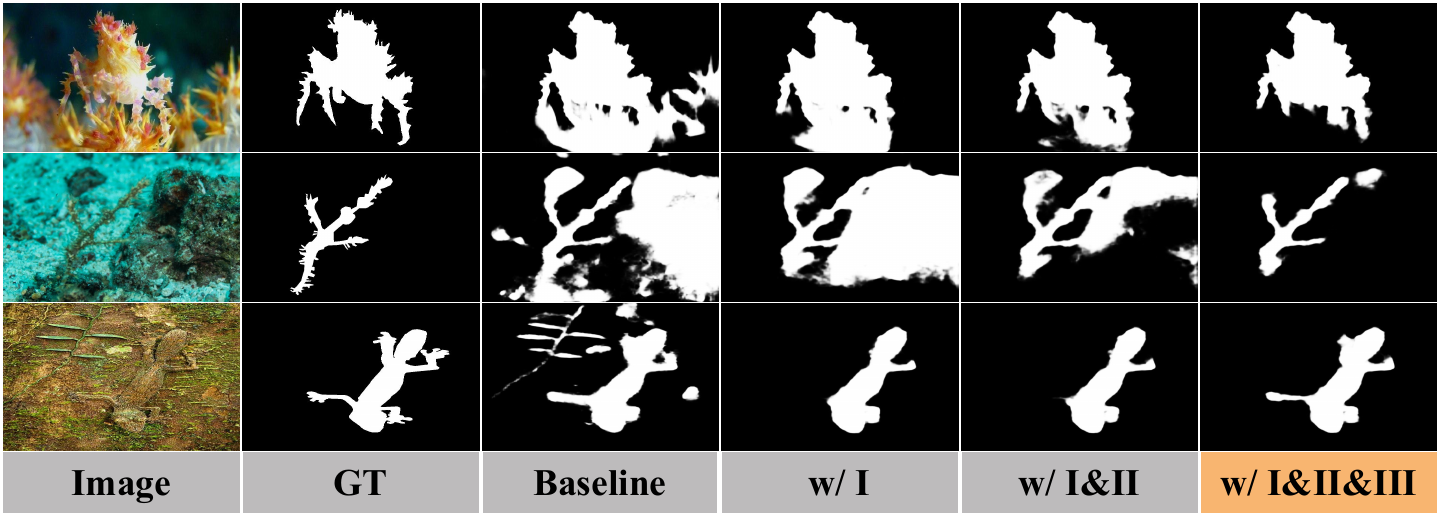}\\%Uni.11.17.5.pdf
\caption{Visualization of the component ablation study. I, II, and III represent prompt adapter, response filter, and semantic matcher, respectively.}
\label{fig:6.5}
% \[]{3mm}
\end{figure}

%During training and inference,the main scale is 384 × 384.
\subsection{Compare with State-of-the-art Methods}
%数字重算！！！
\textbf{Quantitative Comparison.} Being the first WSCOD method to incorporate point, scribble and box supervision, the proposed approach primarily leverages scribble supervision and full (mask) supervision as baselines.
%our average model performance under three weakly supervised labels shows a significant improvement. On average, MAE increased by 36.8%, Sm increased by 12.7%, Em increased by 8.7%, and Fw_beta increased by 21.6%.
As demonstrated in Tab.\ \ref{tab:1}, our method achieves substantial improvements, we averaged the results under three weakly-supervised labels, with an average enhancement of $26.8\%$ for MAE, $6.1\%$ for S$_{m}$, and $5.5\%$ for E$_{m}$ compared to the state-of-the-art weakly-supervised COD method, WS-SAM~\cite{he2024weakly}. 
In particular, our approach performs exceptionally well under point and box supervision. It highlights our capability to achieve better performance with fewer annotations. Our approach even outperforms the state-of-the-art fully supervised method, ZoomNet\ \cite{pang2022zoom}. To verify the advantages of our method over simply using of SAM, we compare with SAM-S and SAM-P, which fine-tune the mask decoder of SAM with scribble and point supervisions, respectively, by the partial cross-entropy loss. When testing, SAM-S and SAM-P use the automatic prompt generation strategy and report the results with the highest IoU scores. We do see performance gains after finetuning SAM with point (SAM-P) and scribble (SAM-S) supervision, but the results are still far below our method. This demonstrates the superiority of our method, which utilizes SAM prompt-adaptive knowledge distillation for small models.
To further analyze the segmentation quality, we draw the density distribution map about S$_{m}$ and object size on the test dataset in Fig.\ \ref{fig:distru}. 
It can be observed that the proposed method achieves an overall improvement and more stable performance on arbitrary sized objects compared to CRNet and ZoomNet. Especially for challenging small and large objects, our model has a significant improvement compared to CRNet and ZoomNet.
Specifically, we design the “Mix” training method, i.e., randomly assigning one type of weakly-supervised label to each image in training. 
It is found that the performance is close to that of box-supervised method, particularly demonstrating a significant performance advantage on the CAMO dataset. 
The diversity of training introduced by mixing different labels is beneficial to learn more complex and rich feature representations comprehensively, capturing the feature at various levels.
%有趣的是，我们尝试了mix的标签方式,即混合三种弱监督标签,并且随机抽取一类作为每张图像的监督信息。发现在CAMO数据集上性能大幅提高,并且和最好的单类标签监督相媲美，这可能是因为不同类型label在PKD中,这种多样性可以激发模型学习更复杂和丰富的特征表示。不同的弱监督标签提供了图像不同层面的信息。混合这些标签可以带来更全面的信息

%In addition, we also plotted the Sm distribution, and in comparison to CRNet, the S$_{m}$ distribution chart demonstrates better quality of the segmentation maps we generated.

\begin{table}[t]
  \begin{center}
  \caption{Ablations study of SAM-COD.}\label{tab:ablations}
  % \[]{-15pt}
  \renewcommand\arraystretch{1.4}
\tabcolsep=0.01cm
    {\scriptsize
\begin{tabular}{ccccc|cccc|cccc|cccc}
\toprule[0.8pt]
\multicolumn{5}{c|}{Settings}&\multicolumn{4}{c|}{Box}&\multicolumn{4}{c|}{Point}&\multicolumn{4}{c}{Scribble}\\
\multicolumn{1}{c}{SAM}&\multicolumn{1}{c}{SKD}&\multicolumn{1}{c}{Filt.}&\multicolumn{1}{c}{Match.}&\multicolumn{1}{c|}{PKD}&MAE$\downarrow$&S$_{m}\uparrow$&E$_{m}\uparrow$&F$^{w}_{\beta}\uparrow$&MAE$\downarrow$&S$_{m}\uparrow$&E$_{m}\uparrow$&F$^{w}_{\beta}\uparrow$&MAE$\downarrow$&S$_{m}\uparrow$&E$_{m}\uparrow$&F$^{w}_{\beta}\uparrow$\\
\toprule[0.8pt]
\checkmark&&&&&0.039&0.792&0.866&0.680&0.056&0.793&0.849&0.663&0.041&0.801&0.864&0.696\\
\checkmark&\checkmark&&&&0.037&0.801&0.874&0.685&0.053&0.800&0.865&0.680&0.038&0.814&0.874&0.701\\
\checkmark&\checkmark&\checkmark&&&0.035&0.817&0.883&0.698&0.036&0.821&0.890&0.708&0.035&0.823&0.890&0.711\\
\checkmark&\checkmark&\checkmark&\checkmark&&0.031&0.831&0.903&0.725&0.032&0.829&0.899&0.720&0.032&0.827&0.901&0.722\\
\hline
\rowcolor{lightgray!40}\checkmark&\checkmark&\checkmark&\checkmark&\checkmark&\textbf{0.028}&\textbf{0.842}&\textbf{0.914}&\textbf{0.745}&\textbf{0.028}&\textbf{0.831}&\textbf{0.901}&\textbf{0.725}&\textbf{0.029}&\textbf{0.831}&\textbf{0.904}&\textbf{0.728}\\

\toprule[0.8pt]
\end{tabular}
}
\end{center}
% \[]{-10pt}
\end{table}

\begin{table}[t]
\begin{minipage}{0.50\linewidth}
\renewcommand\arraystretch{1.3}
\tabcolsep=0.07cm
\scriptsize
\centering
\captionsetup{width=.95\textwidth}
\caption{Effect of the operation in prompt adapter. Discret. represents discrete sampling in the prompt adapter.}
\label{tab:adapter}
\begin{tabular}{c|c|cccc}
\toprule[0.8pt]
\multicolumn{1}{c|}{Zhang-suan}&\multicolumn{1}{c|}{Discret.}&MAE$\downarrow$&S$_{m}\uparrow$&E$_{m}\uparrow$&F$^{w}_{\beta}\uparrow$\\
\toprule[0.8pt]
$\times$&$\times$&0.189&0.591&0.592&0.364\\
\checkmark&$\times$&0.093&0.712&0.751&0.519\\
\hline
\rowcolor{lightgray!40}\checkmark&\checkmark&\textbf{0.038}&\textbf{0.814}&\textbf{0.874}&\textbf{0.701}\\
\toprule[0.8pt]
\end{tabular} 
\end{minipage}\begin{minipage}{0.48\linewidth}  
\renewcommand\arraystretch{1.3}
\tabcolsep=0.07cm
\scriptsize
\centering
\captionsetup{width=.95\textwidth}
\caption{The impact of $\alpha$, $\tau_s$, and $\tau_b$ in model.}\label{tab:hy}
\begin{tabular}{cc|cc|cc}
%缺一个初始掩码的超参数！！！！
\toprule[0.8pt]
$\alpha$ &MAE$\downarrow$&$\tau_s$&MAE$\downarrow$&$\tau_b$&MAE$\downarrow$\\
\toprule[0.8pt]
0.025&0.031&0.001&0.038&0.5&0.032\\
0.050&0.030&0.003&0.036&\textbf{0.6}&\textbf{0.029}\\
\textbf{0.075}&\textbf{0.029}&\textbf{0.005}&\textbf{0.029}&0.7&0.029\\
0.100&0.038&0.010&0.039&0.8&0.033\\
\toprule[0.8pt]
\end{tabular}
\end{minipage}
\end{table}

\begin{table}[t]
\begin{minipage}{0.50\linewidth}
\renewcommand\arraystretch{1.3}
\tabcolsep=0.07cm
\scriptsize
\centering
\captionsetup{width=.97\textwidth}
\caption{The ablation study for different augmentations in Self-Knowledge Distillation. “S”, “C”, “T”, “F”, “G” are Scale, Crop, Translate, Flip, Guassblur, respectively.
%an area generation strategy, respectively. 
}\label{tab:aug}

\begin{tabular}{ccccc|cccc}
\toprule[0.8pt]
\multicolumn{5}{c|}{Augmentations}&\multirow{2}*{MAE$\downarrow$}&\multirow{2}*{S$_{m}\uparrow$}&\multirow{2}*{E$_{m}\uparrow$}&\multirow{2}*{F$^{w}_{\beta}\uparrow$}\\
S&C&T&F&G&&&&\\
%{G}&MAE$\downarrow$&S$_{m}\uparrow$&E$_{m}\uparrow$&F$^{w}_{\beta}\uparrow$\\
\toprule[0.8pt]
&&&&&0.033&0.826&0.897&0.714\\
\checkmark&&&&&0.032&0.829&0.899&0.721\\
\checkmark&\checkmark&&&&0.032&0.828&0.898&0.719\\
\checkmark&\checkmark&\checkmark&&&0.032&0.831&0.901&0.723\\
\checkmark&\checkmark&\checkmark&\checkmark&&0.031&0.830&0.901&0.723\\
\hline
\rowcolor{lightgray!40}\checkmark&\checkmark&\checkmark&\checkmark&\checkmark&\textbf{0.029}&\textbf{0.835}&\textbf{0.906}&\textbf{0.732}\\
\toprule[0.8pt]
\end{tabular} 
\end{minipage}\begin{minipage}{0.48\linewidth}  
\renewcommand\arraystretch{1.3}
\tabcolsep=0.07cm
\scriptsize
\centering
\captionsetup{width=.97\textwidth}
\caption{Ablation study on knowledge distillation losses. MSE, L1, CE mean Mean Square Error, L1, and Cross Entropy loss, respectively.}\label{tab:kdloss}
\begin{tabular}{ll|cccc}
\toprule[0.8pt]
\multicolumn{1}{c}{SKD}&\multicolumn{1}{c|}{PKD}&MAE$\downarrow$&S$_m\uparrow$&E$_m\uparrow$&F$^w_\beta\uparrow$\\
\toprule[0.8pt]
w/o&w/o&0.039&0.801&0.873&0.684\\
\hline
w/MSE &w/o&0.038&0.801&0.878&0.693\\
w/CE &w/o&0.039&0.798&0.875&0.689\\
\rowcolor{lightgray!40}w/L1 &w/o&\textbf{0.037}&\textbf{0.807}&\textbf{0.882}&\textbf{0.701}\\
\hline
w/o&w/ MSE&0.035&0.803&0.840&0.669\\
w/o&w/ L1&\textbf{0.032}&0.788&0.845&0.678\\
\rowcolor{lightgray!40}w/o&w/ CE&0.034&\textbf{0.821}&\textbf{0.891}&\textbf{0.707}\\
\toprule[0.8pt]
\end{tabular}
\end{minipage}
% \[]{-10pt}
\end{table}

% \begin{table}[t]
%   \begin{center}
%     {\footnotesize
% \begin{tabular}{c|c|cccc}
% \hline
% \multicolumn{1}{c}{Zhang-suan}&\multicolumn{1}{|c|}{Discret.}&MAE$\downarrow$&S$_{m}\uparrow$&E$_{m}\uparrow$&F$^{w}_{\beta}\uparrow$\\
% \hline
% \hline
% $\times$&$\times$&0.189&0.591&0.592&0.364\\
% \checkmark&$\times$&0.093&0.712&0.751&0.519\\
% \checkmark&\checkmark&\textbf{0.038}&\textbf{0.814}&\textbf{0.874}&\textbf{0.701}\\
% \hline
% \end{tabular}
% }
% \end{center}
% \caption{Effect of the operation in prompt adapter. Discret. represents discrete sampling in the prompt adapter.}
% \label{tab:adapter}
% \end{table}

% \begin{table}[t]
%   \begin{center}
%   \tabcolsep=0.05cm
%     {\scriptsize
% \begin{tabular}{cc|cc|cc}
% %缺一个初始掩码的超参数！！！！
% \hline
% $\alpha$ &MAE$\downarrow$&$\tau_s$&MAE$\downarrow$&$\tau_b$&MAE$\downarrow$\\
% \hline
% \hline
% 0.025&0.031&0.001&0.038&0.5&0.032\\
% 0.050&0.030&0.003&0.036&\textbf{0.6}&\textbf{0.029}\\
% \textbf{0.075}&\textbf{0.029}&\textbf{0.005}&\textbf{0.029}&0.7&0.029\\
% 0.100&0.038&0.010&0.039&0.8&0.033\\
% \hline
% \end{tabular}
% }
% \end{center}
%  \caption{The impact of $\alpha$, $\tau_s$, and $\tau_b$ in model.}\label{tab:hy}
% \end{table}

%Change:10.22
\noindent\textbf{Qualitative Evaluation.} Our method produces prediction maps characterized by clearer and more complete object regions, along with sharper contours, significantly outperforming state-of-the-art weakly-supervised COD method CRNet\ \cite{he2023weakly} and fully supervised COD method ZoomNet\ \cite{pang2022zoom},  as shown in Fig.\ \ref{fig:5}. Our method performs well in various challenging scenarios, 
including scenarios with tiny objects (row 3), huge objects (row 4), high intrinsic similarities (row 2), indefinable boundaries (row 2 and 3), and complex backgrounds (row 1). 
%\textcolor{blue}{More results are provided in the \textit{Supplementary Material}}.

\noindent\textbf{Parameter Complexity.} Under similar parameter complexity and computational cost overhead, our model outperforms fully-supervised method ZoomNet\ \cite{pang2022zoom}, as shown in Tab.~\ref{tab:2}.

\subsection{Ablation Study} 
As COD10K is the most representative dataset, all following ablation experiments are performed on it. Unless specifically indicated, all results are the averages of three different prompts (point, box, and scribble).

%In this section, we conduct a thorough set of ablation analyses on various components. Given that CAMO represents the most challenging dataset, as indicated by the lowest scores in Tab. \ref{tab:1}, all subsequent ablation experiments are conducted on this dataset.

%In this part, we perform comprehensive ablation analyses on different components. Because CAMO is the most challenging dataset, where the methods obtain the worst scores according to Tab. \ref{tab:1}, all subsequent ablation experiments are carried out on it. 

%Include 

\noindent\textbf{Effectiveness of Prompt Adapter.} The ablation results of prompt adapter are presented in Tab.\ \ref{tab:adapter}. Adapter has a large influence on the performance for scribble prompt. In addition, compared with baseline, a more accurate prediction map can be obtained by using the adapter, as shown in Fig.\ \ref{fig:6.5}. In addition, adapter has a hyperparameter $\alpha$ to control the degree of discrete sampling, as shown in Tab.\ \ref{tab:hy}, with optimal effects achieved for suitable discrete sampling.

\noindent\textbf{Effectiveness of Response Filter.} As shown in Tab.\ \ref{tab:ablations}, the results are significantly improved using response filter. Fig.\ \ref{fig:6.5} intuitively illustrates that the response filter enhances the precision of prediction maps. In addition, response filter has two hyperparameters $\tau_s$ and $\tau_b$ to control effects, as shown in Tab.\ \ref{tab:hy}.

\noindent\textbf{Effectiveness of Semantic Matcher.} We conduct ablation experiments for the semantic matcher, as shown in Tab. \ref{tab:ablations}. In addition, a more complete visualization of the prediction map can be obtained by using semantic matcher, as shown in Fig.\ \ref{fig:6.5}.

\begin{figure}[t]
\centering
\includegraphics[width=11.5cm]{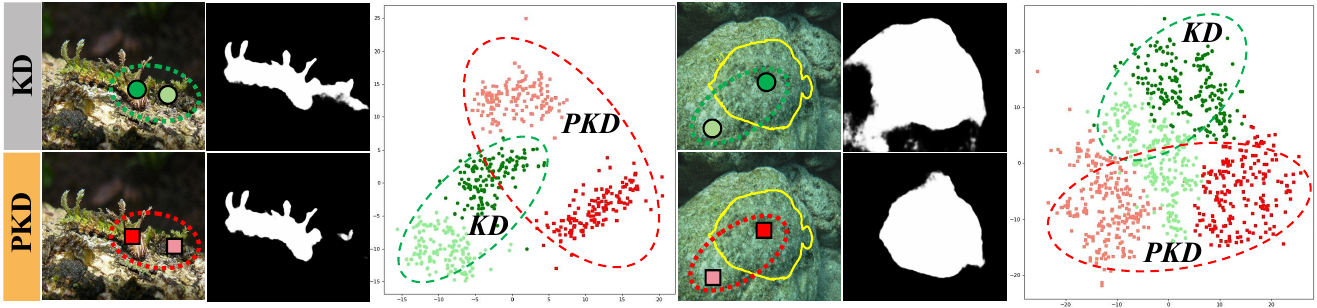}\\
\caption{Visualization of the feature. Entangled features from foregrounds and backgrounds are well separated by our prompt-adaptive KD. (visualized by t-SNE). Green and red colors represent features acquired under KD and PKD, respectively. 
%The light and dark colors represent foreground and background features, respectively.
}
\label{fig:6}
\end{figure}

\noindent\textbf{Effectiveness of Prompt-Adaptive KD.} We test the effect of prompt-adaptive KD compared with traditional KD. Tab.\ \ref{tab:ablations} shows that our PDK has a better performance. Additionally, using PKD also enhances the precision of prediction maps and able to continuously optimize representation and separate entangled object from background, making the model eventually learn robust representations, as shown in Fig.\ \ref{fig:6}. Tab.\ \ref{tab:kdloss} shows that CE loss performs best in PKD.

% \begin{table}[t]
%   \begin{center}
%   \caption{Comparison with state-of-the-art WSSOD methods in SOD task.}\label{tab:10}
%     \tabcolsep=0.10cm
%     {\scriptsize
% \begin{tabular}{c|ccc|cccc}
% \hline
% \multirow{2}*{Methods}&\multirow{2}*{Label}&\multirow{2}*{Para.}&\multirow{2}*{MACs}&\multicolumn{4}{c}{MAE $\downarrow$}\\
% &&&&ECSSD&DUT-O&HKU-IS&DUTS-TE\\
% \hline

% AFNet&F&-&-&0.0418&0.0574&0.0358&0.0458\\
% BASNet&F&-&-&0.0370&0.0565&0.0322&0.0476\\
% GateNet&F&-&-&0.0401&0.0549&0.0331&0.0401\\
% \hline
% SCWSSOD&S&-&-&0.0489&\textbf{\color{blue}0.0602}&0.0375&0.0487\\
% PSOD&P&104.0&107.9&\textbf{\color{blue}0.0358}&0.0643&\textbf{\color{blue}0.0322}&\textbf{\color{blue}0.0449}\\
% %PCOD (ours)&P&62.6&52.6&0.0451&0.0599&0.1187&0.0311&0.0422\\ 
% \hdashline
% Ours&P&62.6&52.6&0.0326&0.0512&0.0242&0.0342\\
% Ours&S&62.6&52.6&0.0335&0.0505&0.0244&0.0327\\
% Ours&B&62.6&52.6&0.0315&0.0507&0.0233&0.0334\\
% Ours&W&62.6&52.6&\textbf{\color{red}0.0325}&\textbf{\color{red}0.0508}&\textbf{\color{red}0.0239}&\textbf{\color{red}0.0334}\\
% \hline
% \end{tabular}
% }
% \end{center}
% \end{table}

\begin{table*}[!t]
  \begin{center}
   \caption{Comparison with state-of-the-art WSSOD methods in SOD task.}\label{tab:10}
   % \[]{-10pt}
   \renewcommand\arraystretch{1.3}
  \tabcolsep=0.01cm
    {\scriptsize
\begin{tabular}{l|c|ccc|ccc|ccc|ccc}
\toprule[0.8pt]
\multirow{2}*{Methods}&\multirow{2}*{Label}&\multicolumn{3}{c|}{ECSSD}&\multicolumn{3}{c|}{DUT-O}&\multicolumn{3}{c|}{HKU-IS} &\multicolumn{3}{c}{DUTS-TE}\\%&\multicolumn{2}{|c}{ILSVRC2012}\\

&&MAE $\downarrow$&S$_{m} \uparrow$&F$^{max}_{\beta} \uparrow$&MAE $\downarrow$&S$_{m} \uparrow$&F$^{max}_{\beta} \uparrow$&MAE $\downarrow$&S$_{m} \uparrow$&F$^{max}_{\beta} \uparrow$&MAE $\downarrow$&S$_{m} \uparrow$&F$^{max}_{\beta} \uparrow$\\
\toprule[0.8pt]

AFNet~\cite{feng2019attentive}&F& 0.042&0.913&0.935 &0.057&0.826 &0.797&0.036&0.905&0.923&0.046&0.867 &0.863 \\
BASNet~\cite{qin2019basnet}&F&0.037&0.916&0.943&0.057&0.836&0.805&0.032&0.909&0.928&0.048&0.866&0.859\\
\hline
SCSOD~\cite{yu2021structure}&S&0.049&0.881&0.914&0.060&0.811&0.782&0.038&0.882&0.908&0.049&0.853&0.858\\
PSOD~\cite{gao2022weakly}&P&0.036&0.913&0.935&0.064&0.824&0.808&0.033&0.901&0.923&0.045&0.853&0.858\\

\hline
\rowcolor{lightgray!40}SAM-COD&P&0.033&0.925&0.947&0.051&0.844&0.826&0.024&0.946&0.941&0.034&0.892&0.898\\
\rowcolor{lightgray!40}SAM-COD&S&0.034&0.921&0.944&\textbf{0.050}&\textbf{0.846}&\textbf{0.829}&0.024&0.947&0.944&\textbf{0.033}&0.898&0.901\\
\rowcolor{lightgray!40}SAM-COD&B&\textbf{0.031}&\textbf{0.929}&\textbf{0.952}&0.051&0.844&0.828&\textbf{0.023}&\textbf{0.952}&\textbf{0.949}&\textbf{0.033}&\textbf{0.899}&\textbf{0.903}\\
\toprule[0.8pt]
\end{tabular}
}
\end{center}
% \[]{-10pt}
\end{table*}

\noindent\textbf{Effectiveness of Self-Knowledge Distillation.} We conduct ablation experiments for SKD. Firstly, we separately test the performance of models with and without SKD, our proposed self-knowledge distillation obtains a significant improvement, as shown in Tab.\ \ref{tab:ablations}. In addition, we conduct exhaustive experiments for data augmentation, which is an important operation for SKD, as shown in Tab.\ \ref{tab:aug}. We test different types of knowledge distillation losses and find that L1 loss is performing best, as shown in Tab.\ \ref{tab:kdloss}.

%We conduct ablation experiments for self knowledge distillation. Firstly, we separately tested the performance of models with and without SKD, our proposed self-knowledge distillation obtains a significant improvement, as shown in Tab. \ref{tab:SKD}. In addition, we conduct exhaustive experiments for data augmentation, which is important for SKD. Specifically, the augmentation is divided into three aspects, color-texture related (GuassBlur), position related (Flip, Translate), and size related (Crop, Scale). It shows that color-texture and size augmentation can improve the performance, but position augmentation yields unsatisfactory results, as shown in Tab. \ref{tab:aug}. Additionally, using self-knowledge distillation enhances the precision of prediction maps, as shown in Fig. \ref{fig:6}, which also show that model self knowledge-distillation is able to continuously optimize representation and separate entangled object from background, making the model eventually learn robust representations.

%\subsubsection{Impact of the Hyperparameter \textbf{$\alpha$, $\tau$, $\lambda$, $w$.}} $\alpha$ represents the balance factor between Self Knowledge Distillation and Knowledge Distillation.

% \noindent\textbf{Effectiveness of KD Loss Functions.} As shown in Table \ref{tab:kdloss}, we divided the model into two parts: prompt-adapter knowledge distillation and self knowledge distillation. We tested different loss types for each part and found that CE loss and L1 loss performed the best, respectively. 

% \\[4pt]
\subsection{Extension to SOD}
%%%%11.14:放到附加材料 / 不放%%%%
% To demonstrate the strong transferability of our model, we conducted the following experiments:

% \textbf{Extension to other WSCOD method}. We apply the proposed model to other weakly supervised COD
% models, i.e., scribble supervised COD method CRNet\cite{he2023weakly}, as shown in Table \ref{tab:9}, which demonstrates the generality of the proposed method.

% \textbf{Extension to COD methods.} Our model is relatively simple, especially in the design of the encoder and decoder. We haven't incorporated overly complex structures, which can be replaced with various types of existing network models according to the requirements, such as lightweight, high-performance, or a compromise between the two, making our model quite flexible.  As shown in Tab. \ref{tab:9}, Our method can be adapted to various types of COD and WSCOD models. When transferred to weakly supervised models, our approach significantly improves performance under weak supervision. When transferred to fully supervised models, we can achieve acceptable performance degradation using lower-level labels compared to full supervision.

% \textbf{Extension to SOD.} 
Our method excels not only in COD but also demonstrates remarkable performance in SOD. Specifically, we train on the SOD dataset using the labels of point, scribble, and box, respectively, and the results obtained are shown in Tab.\ \ref{tab:10}. 
%(More results in supplementary materials).
% In addition, We also apply our method to Polyp image segmentation, as shown in supplementary materials. 
We attribute this success to our exploration of the potential of SAM and improvements in knowledge distillation. which contributes to our strong performance in WSSOD. 

%%%% 11.14:放到附加材料%%%%
% \subsection{Limitations}
% As shown in Fig. \ref{fig:Failure}, we visualized some of the failed cases, primarily due to incomplete segmentation and missegmentation. This is mainly attributed to xxx. Therefore, future work needs to consider more the interaction between object and context to overcome the problems of object occlusion and detection more.

% \begin{figure}[t]
% \centering
% \includegraphics[width=8.5cm]{sec/Fig/Filure.png}\\
% \caption{Two failure cases of our method.}
% \label{fig:Failure}
% \end{figure}

% \section{Future Work}
% This study showcases the effectiveness of our proposal model in COD and SOD. Moving forward, we plan to extend the model to tackle other challenging image segmentation tasks (such as medical image segmentation, shadow detection) and broaden its application to other fields. 
%%%% 11.14:放到附加材料%%%%

\subsection{Discussion}
\textbf{Is the prompt-adaptive KD from SAM important?}

\begin{figure}[t]
\centering
\includegraphics[width=12cm]{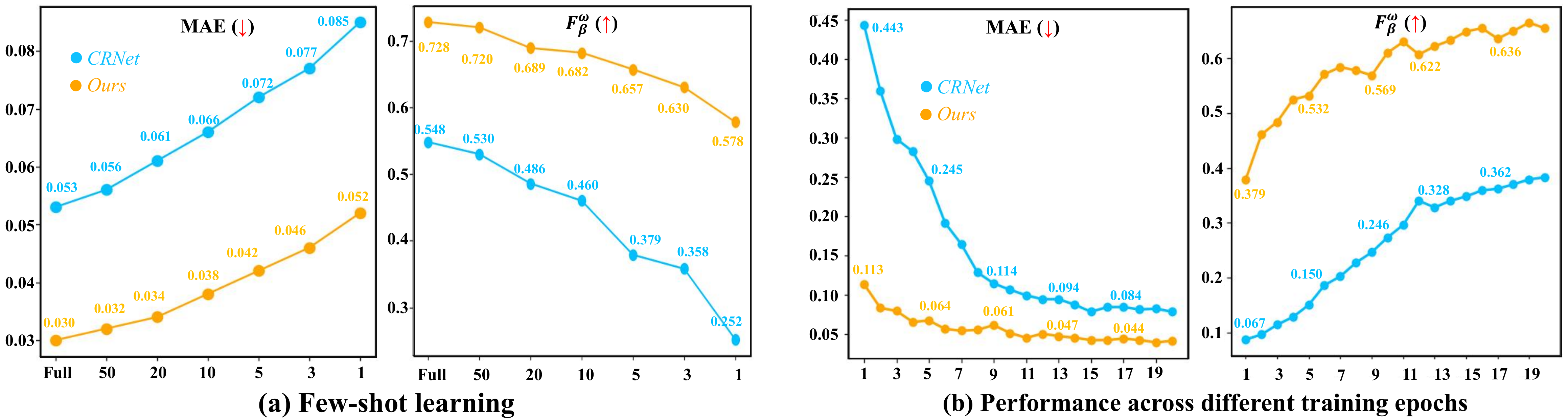}\\
% \[]{-5pt}
\caption{
%The benefits of distilling knowledge from SAM. 
The benefits of prompt-adaptive distilling knowledge from SAM. 
(a) Data efficiency: Few-shot performance. 
%Figure plot of the $k$-shot segmentation performance. 
For each $k$-shot setting, we repeat the experiment $5$ times to randomly select $k$ images as training data. The average results are shown in the curve. (b) Training efficiency: Performance across different training epochs with the same training setting. }
\label{fig:8}
% \[]{-10pt}
\end{figure}

\noindent\textbf{1) Data efficiency.} 
We also evaluate the performance of our model and CRNet in few-shot setting, as shown in Fig.\ \ref{fig:8}(a). 
%Specifically, our model are retained only using the COD10K-Train dataset that which contains categories, and test in COD10K-Test dataset.
Specifically, our model is trained only using the COD10K-Train dataset, which contains categories, and tested on the COD10K-Test dataset.
Compared to CRNet, our model achieves promising results with much fewer training data.
Especially in the extreme scenario, our method only uses one training image in each category, performance significantly surpasses that of CRNet training on the complete dataset. 
Fig.\ \ref{fig:8}(a) verifies the effectiveness and efficiency of the proposed method.
%It can be attributed to the use of SAM, which is trained on a large amount of data and handles abundant segmentation knowledge. 
Through prompt-adaptive knowledge distillation, we transfer the knowledge from SAM to our model, 
%It enables our model to achieve promising results, 
only requiring a small amount of data. 
%In contrast, CRNet requires large amounts of training data to serve on the COD task.

\noindent\textbf{2) Training efficiency.} 
We visualize the curves of various metrics during the training, as shown in Fig.\ \ref{fig:8}(b), 
where CRNet and our model share the same implementation details, including the optimizer, learning rate, epochs, and other relevant parameters. It is observed that our model demonstrates extremely fast convergence speed. To achieve the same performance, our model only needs one epoch of training, while CRNet typically 
requires more than 10 epochs. 
%\textcolor{red}{Our model demonstrates faster training stability, as evidenced by its quicker convergence to smooth curves.} 
Because our model transfers the teacher knowledge from SAM to our small model through prompt-Adaptive knowledge distillation, which is much faster than learning a model from scratch.

% \begin{figure}[t]
% \centering
% \includegraphics[width=12cm]{sec/Fig/Uni.11.17.9.pdf}\\
% \caption{Performance across different training epochs with the same training setting.}
% \label{fig:7}
% \end{figure}

% \[]{-5pt}
\section{Conclusion} 
% \[]{-5pt}
In this paper, we propose a SAM-guided unified framework for weakly-supervised camouflaged object detection (WSCOD), named SAM-COD. It integrates all the existing labels for camouflaged objects (\emph{i.e.}, scribbles, bounding boxes, and points), and achieves remarkable performance against the state-of-the-art weakly-supervised methods and even fully-supervised methods. The proposed SAM-COD typically aims to address the issues of SAM in the WSCOD task, \emph{i.e.}, prompt compatibility of the scribble labels, extreme response, semantically erroneous response, and unstable feature representations. Specifically, in SAM-COD, we devise a prompt adapter to handle different labels and employ response filter and semantic matcher to mitigate the issue of imperfect outputs of SAM for camouflaged objects. Moreover, a prompt-adaptive knowledge distillation is proposed for reliable feature representations. We have conducted extensive experiments on camouflaged object datasets, demonstrating the effectiveness of the proposed method, which improves SAM to be more applicable to 
WSCOD.

\bibliographystyle{splncs04}
\bibliography{main}

\begin{thebibliography}{10}
\providecommand{\url}[1]{\texttt{#1}}
\providecommand{\urlprefix}{URL }
\providecommand{\doi}[1]{https://doi.org/#1}

\bibitem{buciluǎ2006model}
Buciluǎ, C., Caruana, R., Niculescu-Mizil, A.: Model compression. In: Proceedings of the 12th ACM SIGKDD international conference on Knowledge discovery and data mining. pp. 535--541 (2006)

\bibitem{caron2021emerging}
Caron, M., Touvron, H., Misra, I., J{\'e}gou, H., Mairal, J., Bojanowski, P., Joulin, A.: Emerging properties in self-supervised vision transformers. In: Proceedings of the IEEE/CVF international conference on computer vision. pp. 9650--9660 (2021)

\bibitem{chen2023sam}
Chen, T., Zhu, L., Ding, C., Cao, R., Zhang, S., Wang, Y., Li, Z., Sun, L., Mao, P., Zang, Y.: Sam fails to segment anything?--sam-adapter: Adapting sam in underperformed scenes: Camouflage, shadow, and more. arXiv preprint arXiv:2304.09148  (2023)

\bibitem{chen2021exploring}
Chen, X., He, K.: Exploring simple siamese representation learning. In: Proceedings of the IEEE/CVF conference on computer vision and pattern recognition. pp. 15750--15758 (2021)

\bibitem{fan2017structure}
Fan, D.P., Cheng, M.M., Liu, Y., Li, T., Borji, A.: Structure-measure: A new way to evaluate foreground maps. In: Proceedings of the IEEE international conference on computer vision. pp. 4548--4557 (2017)

\bibitem{ijcai2018p97}
Fan, D.P., Gong, C., Cao, Y., Ren, B., Cheng, M.M., Borji, A.: Enhanced-alignment measure for binary foreground map evaluation. In: Proceedings of the Twenty-Seventh International Joint Conference on Artificial Intelligence, {IJCAI-18}. pp. 698--704 (2018)

\bibitem{fan2021concealed}
Fan, D.P., Ji, G.P., Cheng, M.M., Shao, L.: Concealed object detection. IEEE transactions on pattern analysis and machine intelligence  \textbf{44}(10),  6024--6042 (2021)

\bibitem{fan2020camouflaged}
Fan, D.P., Ji, G.P., Sun, G., Cheng, M.M., Shen, J., Shao, L.: Camouflaged object detection. In: Proceedings of the IEEE/CVF conference on computer vision and pattern recognition. pp. 2777--2787 (2020)

\bibitem{fan2020pranet}
Fan, D.P., Ji, G.P., Zhou, T., Chen, G., Fu, H., Shen, J., Shao, L.: Pranet: Parallel reverse attention network for polyp segmentation. In: International conference on medical image computing and computer-assisted intervention. pp. 263--273. Springer (2020)

\bibitem{fan2020inf}
Fan, D.P., Zhou, T., Ji, G.P., Zhou, Y., Chen, G., Fu, H., Shen, J., Shao, L.: Inf-net: Automatic covid-19 lung infection segmentation from ct images. IEEE transactions on medical imaging  \textbf{39}(8),  2626--2637 (2020)

\bibitem{feng2019attentive}
Feng, M., Lu, H., Ding, E.: Attentive feedback network for boundary-aware salient object detection. In: Proceedings of the IEEE/CVF conference on computer vision and pattern recognition. pp. 1623--1632 (2019)

\bibitem{perez2012early}
P{\'e}rez-de~la Fuente, R., Delcl{\`o}s, X., Pe{\~n}alver, E., Speranza, M., Wierzchos, J., Ascaso, C., Engel, M.S.: Early evolution and ecology of camouflage in insects. Proceedings of the National Academy of Sciences  \textbf{109}(52),  21414--21419 (2012)

\bibitem{gao2022weakly}
Gao, S., Zhang, W., Wang, Y., Guo, Q., Zhang, C., He, Y., Zhang, W.: Weakly-supervised salient object detection using point supervision. In: Proceedings of the AAAI Conference on Artificial Intelligence. pp. 670--678 (2022)

\bibitem{he2024weakly}
He, C., Li, K., Zhang, Y., Xu, G., Tang, L., Zhang, Y., Guo, Z., Li, X.: Weakly-supervised concealed object segmentation with sam-based pseudo labeling and multi-scale feature grouping. Advances in Neural Information Processing Systems  \textbf{36} (2024)

\bibitem{he2023weakly}
He, R., Dong, Q., Lin, J., Lau, R.W.: Weakly-supervised camouflaged object detection with scribble annotations. In: Proceedings of the AAAI Conference on Artificial Intelligence. pp. 781--789 (2023)

\bibitem{hinton2015distilling}
Hinton, G., Vinyals, O., Dean, J.: Distilling the knowledge in a neural network. stat  \textbf{1050}, ~9 (2015)

\bibitem{ji2021progressively}
Ji, G.P., Chou, Y.C., Fan, D.P., Chen, G., Fu, H., Jha, D., Shao, L.: Progressively normalized self-attention network for video polyp segmentation. In: International Conference on Medical Image Computing and Computer-Assisted Intervention. pp. 142--152. Springer (2021)

\bibitem{ji2021learning}
Ji, W., Yu, S., Wu, J., Ma, K., Bian, C., Bi, Q., Li, J., Liu, H., Cheng, L., Zheng, Y.: Learning calibrated medical image segmentation via multi-rater agreement modeling. In: Proceedings of the IEEE/CVF Conference on Computer Vision and Pattern Recognition. pp. 12341--12351 (2021)

\bibitem{kirillov2023segment}
Kirillov, A., Mintun, E., Ravi, N., Mao, H., Rolland, C., Gustafson, L., Xiao, T., Whitehead, S., Berg, A.C., Lo, W.Y., Doll{\'a}r, P., Girshick, R.B.: Segment anything. 2023 IEEE/CVF International Conference on Computer Vision  (2023)

\bibitem{le2019anabranch}
Le, T.N., Nguyen, T.V., Nie, Z., Tran, M.T., Sugimoto, A.: Anabranch network for camouflaged object segmentation. Computer vision and image understanding  \textbf{184},  45--56 (2019)

\bibitem{li2021uncertainty}
Li, A., Zhang, J., Lv, Y., Liu, B., Zhang, T., Dai, Y.: Uncertainty-aware joint salient object and camouflaged object detection. In: Proceedings of the IEEE/CVF Conference on Computer Vision and Pattern Recognition. pp. 10071--10081 (2021)

\bibitem{lv2021simultaneously}
Lv, Y., Zhang, J., Dai, Y., Li, A., Liu, B., Barnes, N., Fan, D.P.: Simultaneously localize, segment and rank the camouflaged objects. In: Proceedings of the IEEE/CVF Conference on Computer Vision and Pattern Recognition. pp. 11591--11601 (2021)

\bibitem{margolin2014evaluate}
Margolin, R., Zelnik-Manor, L., Tal, A.: How to evaluate foreground maps? In: Proceedings of the IEEE conference on computer vision and pattern recognition. pp. 248--255 (2014)

\bibitem{mei2021camouflaged}
Mei, H., Ji, G.P., Wei, Z., Yang, X., Wei, X., Fan, D.P.: Camouflaged object segmentation with distraction mining. In: Proceedings of the IEEE/CVF Conference on Computer Vision and Pattern Recognition. pp. 8772--8781 (2021)

\bibitem{pang2022zoom}
Pang, Y., Zhao, X., Xiang, T.Z., Zhang, L., Lu, H.: Zoom in and out: A mixed-scale triplet network for camouflaged object detection. In: Proceedings of the IEEE/CVF Conference on computer vision and pattern recognition. pp. 2160--2170 (2022)

\bibitem{qin2019basnet}
Qin, X., Zhang, Z., Huang, C., Gao, C., Dehghan, M., Jagersand, M.: Basnet: Boundary-aware salient object detection. In: Proceedings of the IEEE/CVF conference on computer vision and pattern recognition. pp. 7479--7489 (2019)

\bibitem{sun2021context}
Sun, Y., Chen, G., Zhou, T., Zhang, Y., Liu, N.: Context-aware cross-level fusion network for camouflaged object detection. In: Proceedings of the Thirtieth International Joint Conference on Artificial Intelligence, {IJCAI-21}. pp. 1025--1031 (2021)

\bibitem{tang2023can}
Tang, L., Xiao, H., Li, B.: Can sam segment anything? when sam meets camouflaged object detection. arXiv preprint arXiv:2304.04709  (2023)

\bibitem{wang2021pyramid}
Wang, W., Xie, E., Li, X., Fan, D.P., Song, K., Liang, D., Lu, T., Luo, P., Shao, L.: Pyramid vision transformer: A versatile backbone for dense prediction without convolutions. In: Proceedings of the IEEE/CVF international conference on computer vision. pp. 568--578 (2021)

\bibitem{yang2021uncertainty}
Yang, F., Zhai, Q., Li, X., Huang, R., Luo, A., Cheng, H., Fan, D.P.: Uncertainty-guided transformer reasoning for camouflaged object detection. In: Proceedings of the IEEE/CVF International Conference on Computer Vision. pp. 4146--4155 (2021)

\bibitem{yu2021structure}
Yu, S., Zhang, B., Xiao, J., Lim, E.G.: Structure-consistent weakly supervised salient object detection with local saliency coherence. In: Proceedings of the AAAI conference on artificial intelligence. pp. 3234--3242 (2021)

\bibitem{zhai2021mutual}
Zhai, Q., Li, X., Yang, F., Chen, C., Cheng, H., Fan, D.P.: Mutual graph learning for camouflaged object detection. In: Proceedings of the IEEE/CVF Conference on Computer Vision and Pattern Recognition. pp. 12997--13007 (2021)

\bibitem{zhan2023camouflage}
ZHAN, C., WANG, A., WANG, M.: Camouflage object segmentation method based on channel attention and edge fusion. Journal of Computer Applications  \textbf{43}(7), ~2166 (2023)

\bibitem{zhang2020weakly}
Zhang, J., Yu, X., Li, A., Song, P., Liu, B., Dai, Y.: Weakly-supervised salient object detection via scribble annotations. In: Proceedings of the IEEE/CVF conference on computer vision and pattern recognition. pp. 12546--12555 (2020)

\bibitem{zhang1984fast}
Zhang, T.Y., Suen, C.Y.: A fast parallel algorithm for thinning digital patterns. Communications of the ACM  \textbf{27}(3),  236--239 (1984)

\end{thebibliography}
\end{document}